\documentclass[journal]{IEEEtran}
\usepackage{graphicx}
\usepackage{multirow}
\usepackage{soul}
\usepackage[table]{xcolor}
\definecolor{softy}{cmyk}{0.1,0.09,0.08,0.0}

\newcommand{\an}[1]{{\textcolor{black}{#1}}}

\ifCLASSINFOpdf
\else
\fi

\begin{document}

\title{Role of Data Augmentation Strategies in \\
Knowledge Distillation for Wearable Sensor Data\\ }

\author{Eun~Som~Jeon,~\IEEEmembership{Student~Member,~IEEE,}
        Anirudh~Som,
        Ankita~Shukla,
        Kristina Hasanaj,
        Matthew~P.~Buman,
        and~Pavan~Turaga,~\IEEEmembership{Senior~Member,~IEEE}

\thanks{E. Jeon, A. Shukla and P. Turaga are with the School of Arts, Media and Engineering and School of Electrical, Computer and Energy Engineering, Arizona State University, Tempe, AZ 85281 USA email: (ejeon6@asu.edu; Ankita.Shukla@asu.edu; pturaga@asu.edu).}
\thanks{A. Som is with the Center for Vision Technologies Group at SRI International, Princeton, NJ 08540 USA email: Anirudh.Som@sri.com}
\thanks{ K. Hasanaj and M. P. Buman are with the College of Health Solutions, Arizona State University, Phoenix, AZ 85004 USA email: (khasanaj@asu.edu; mbuman@asu.edu).
}
\thanks{This has been accepted at the IEEE Internet of Things Journal.}
}

\markboth{IEEE INTERNET OF THINGS JOURNAL,~Vol.~0, No.~0, January~2022}%
{Jeon \MakeLowercase{\textit{et al.}}: Draft of IEEEtran.cls for IEEE Journals}

\maketitle


\begin{abstract}
Deep neural networks are parametrized by several thousands or millions of parameters, and have shown tremendous success in many classification problems. However, the large number of parameters makes it difficult to integrate these models into edge devices such as smartphones and wearable devices. To address this problem, knowledge distillation (KD) has been widely employed, that uses a pre-trained high capacity network to train a much smaller network, suitable for edge devices. In this paper, for the first time, we study the applicability and challenges of using KD for time-series data for wearable devices. Successful application of KD requires specific choices of data augmentation methods during training. However, it is not yet known if there exists a coherent strategy for choosing an augmentation approach during KD.  In this paper, we report the results of a detailed study that compares and contrasts various common choices and some hybrid data augmentation strategies in KD based human activity analysis. Research in this area is often limited as there are not many comprehensive databases available in the public domain from wearable devices. Our study considers databases from small scale publicly available to one derived from a large scale interventional study into human activity and sedentary behavior. We find that the choice of data augmentation techniques during KD have a variable level of impact on end performance, and find that the optimal network choice as well as data augmentation strategies are specific to a dataset at hand. However, we also conclude with a general set of recommendations that can provide a strong baseline performance across databases.

\end{abstract}


\IEEEpeerreviewmaketitle
\begin{figure*}[t!]
\includegraphics[scale=0.62] {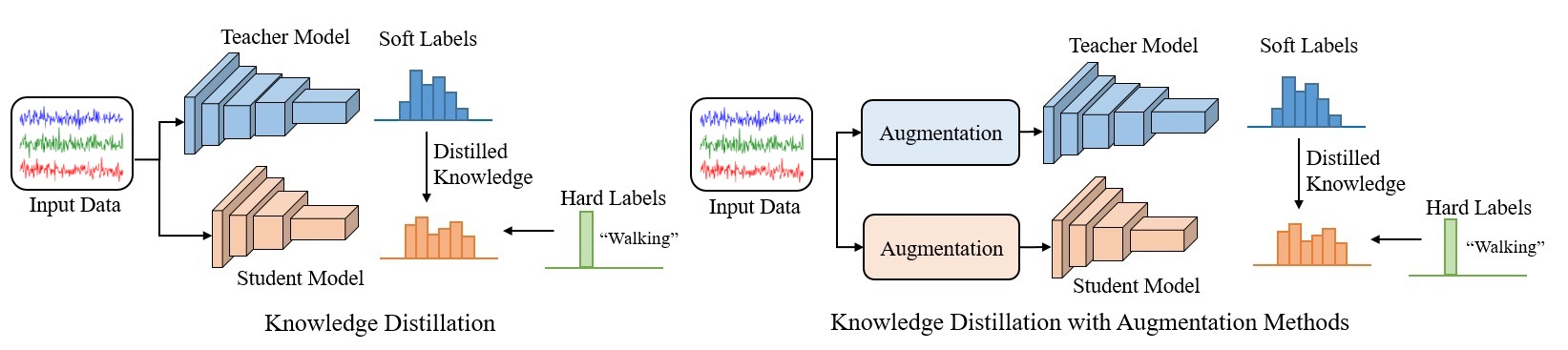}
\centering
\caption{An overview of standard knowledge distillation framework (left) and proposed knowledge distillation with data augmentation method (right). A high capacity network known as teacher is used to guide the learning of a smaller network known as student. A set of augmentations strategies are used to train both the teacher and student networks.}
\label{figure:KD}
\end{figure*}

\section{Introduction}
\IEEEPARstart{D}{eep Learning} has achieved state-of-the-art performance in various fields, including computer vision \cite{he2016deep,huang2017densely, dalal2005histograms,lowe2004distinctive}, speech recognition \cite{abdel2014convolutional}, \cite{xiong2018microsoft}, and wearable sensors analysis \cite{wan2020deep}, \cite{fawaz2019deep}. In general, stacking more layers or increasing the number of learnable parameters causes deep networks to exhibit improved performance \cite{huang2017densely}, \cite{dalal2005histograms}, \cite{lowe2004distinctive}, \cite{fawaz2019deep}, \cite{khan2020survey}, \cite{gil2020improving}.
However, this causes the model to become large resulting in additional need for compute and power resources, for training, storage, and deployment. These challenges can hinder the ability to incorporate such models into edge devices.
Many studies have explored techniques such as network pruning \cite{molchanov2016pruning}, \cite{han2015deep}, quantization \cite{han2015deep}, \cite{wu2016quantized}, low-rank factorization \cite{tai2015convolutional}, and Knowledge Distillation (KD) \cite{hinton2015distilling} to compress deep learning models.
At the cost of lower classification accuracy, some of these methods help to make the deep learning model smaller and increase the speed of inference on the edge devices. Post-training or fine-tuning strategies can be applied to recover the lost classification performance \cite{han2015deep}, \cite{wu2016quantized}. On the contrary, KD does not require fine-tuning nor is subjected to any post-training processes.

KD is a simple and popular technique that is used to develop smaller and efficient models by distilling the learnt knowledge/weights from a larger and more complex model. The smaller and larger models are referred to as student and teacher models, respectively. KD allows the student model to retain the classification performance of the larger teacher model. Recently, different variants of KD have been proposed \cite{yim2017gift}, \cite{heo2019knowledge}. These variations rely on different choices of network architectures, teacher models, and various features used to train the student model. Alongside, teacher models trained by early stopping for KD (ESKD) have been explored, which have helped improving the efficacy of KD \cite{cho2019efficacy}. However, to the best of our knowledge, there is no previous study that explores the effects, challenges, and benefits of KD for human activity recognition using wearable sensor data.

In this paper, we firstly study KD for human activity recognition from time-series data collected from wearable sensors.
Secondly, we also evaluate the role of data augmentation techniques in KD.
This is evaluated by using several time domain data augmentation strategies for training as well as for testing phase. The key highlights and findings from our study are summarized below:
\begin{itemize}
    \item We compare and contrast several KD approaches for time-series data and conclude that EKSD performs better as compared to other techniques.
    \item We perform KD on time-series data with different sizes of teacher and student networks. We corroborate results from previous studies that suggest that the performance of a higher capacity teacher model is not necessarily better.
    \item We study the effects of data augmentation methods on both teacher and student models. We do this to identify which combination of augmentation methods give the most benefit in terms of classification performance.
    \item Our study is evaluated on human activity recognition task and is conducted on a small scale publicly available dataset as well as a large scale dataset. This ensures the observations are reliable irrespective of the dataset sizes.
    
\end{itemize}

The rest of the paper is organized as follows. In Section II, we provide a brief overview of KD techniques as well as data augmentation strategies. In Section III, we present which augmentation methods are used and its effects on time-series data. In Section IV, we describe our experimental results and analysis. In Section V, we discuss our findings and conclusions.

\section{Background} 

\subsubsection{Knowledge Distillation}
\an{The goal of KD is to supervise a small student network by a large teacher network, such that the student network achieves comparable or improved performance over teacher model. This idea was firstly explored by Buciluǎ \emph{et al.} \cite{bucilua2006model} followed by several developments like Hinton \emph{et al.} \cite{hinton2015distilling}.}
The main idea of KD is to use the soft labels which are outputs, soft probabilities, of a trained teacher network and contain more information than just a class label, which is illustrated in Fig. \ref{figure:KD}. For instance, if two classes have high probabilities for a data, the data has to lie close to a decision boundary between these two classes. Therefore, mimicking these probabilities helps student models to get knowledge of teachers that have been trained with labeled data (hard labels) alone.

During training, the loss function $\mathcal{L}$ for a student network is  defined as:
\begin{equation}
\label{eq: stu_loss}
\mathcal{L} = (1-\lambda)\mathcal{L_C} + \lambda \mathcal{L_K}\\
\end{equation}
where $\mathcal{L_C}$ is the standard cross entropy loss, $\mathcal{L_K}$ is KD loss, and $\lambda$ is hyper-parameter; $0 < \lambda < 1$.

In supervised learning, the error between \an{the output of the softmax layer of a student network }and ground-truth label is penalized by the cross-entropy loss:
\begin{equation}
\label{eq: kd_loss}
\mathcal{L_C} = \mathcal{H}(softmax(a_s),y_g)\\
\end{equation}
where $\mathcal{H(\cdot)}$ denotes a cross entropy loss function, $a_s$ is logits of a student (inputs to the final softmax), and $y_g$ is a ground truth label.
In the process of KD, instead of using peaky probability distributions which may produce less accurate results, Hinton \emph{et al.} \cite{hinton2015distilling} proposed to use probabilities with temperature scaling, \emph{i.e.}, output of a teacher network \an{given by} $f_t = softmax(a_t/\tau)$ and a student $f_s = softmax(a_s/\tau)$ are softened by hyperparameter $\tau$, where $\tau > 1$. The teacher and student try to match these probabilities by a KL-divergence loss:
\begin{equation}
\label{eq: kl_loss}
\mathcal{L_K} = \tau^2 KL(f_t,f_s)\\
\end{equation}
where $KL(\cdot)$ is the KL-divergence loss function.

There has been lots of approaches to improve the performance of distillation. Previous methods focus on adding more losses on intermediate layers of a student network to be closer to a teacher \cite{zagoruyko2016paying}, \cite{tung2019similarity}. Averaging consecutive student models tends to produce better performance of students \cite{tarvainen2017mean}. By implementing KD repetitively, the performance of KD is improved, which is called sequential knowledge distillation \cite{zhang2018deep}. 

Recently, learning procedures for improved efficacy of KD has been presented. Goldblum \emph{et al.} \cite{goldblum2020adversarially} suggested adversarially robust distillation (ARD) loss function by minimizing dependencies between output features of a teacher. The method used perturbed data as adversarial data to train the student network. Interestingly, ARD students even show higher accuracy than their teacher.
We adopt augmentation methods to create data which is similar to adversarial data of ARD. Based on ARD, the effect of using adversarial data for KD can be verified, however, \an{which data augmentation is useful for training KD is not well explored}. Unlike ARD, to figure out the role of augmentation methods for KD and which method improves the performance of KD, we use augmentation methods generating different kinds of transformed data for teachers and students. In detail, by adopting augmentation methods, we can generate various combinations of teachers and students which are trained with the same or different augmentation method. It provides to understand which transformation and combinations can improve the performance of KD. We explain the augmentation method for KD in Section III with details.
Additionally, KD tends to show an efficacy with transferring information from early stopped model of a teacher, where training strategy is called ESKD \cite{cho2019efficacy}. Early stopped teachers produce better students than the standard knowledge distillation (Full KD) using fully-trained teachers. Cho \textit{et al.} \cite{cho2019efficacy} presented the efficacy of ESKD with image datasets. We implement ESKD on time-series data and investigate its efficacy on training with data transformed by various augmentation methods. We explain more details in Section III and discuss the efficiency of ESKD in later sections.

In general, many studies focus on the structure of networks and adding loss functions to existing framework of KD \cite{furlanello2018born}, \cite{yang2019training}. However, the performance of most approaches depends on the capacity of student models. Also, availability of sufficient training data for teacher and student models can affect to the final result. In this regard, the factors that have an affect on the distillation process need to be systematically explored, especially on time-series data from wearable sensors.

\subsubsection{Data Augmentation}
Data augmentation methods have been used to boost the generalizability of models and avoid over-fitting. They have  been used in many applications such as time-series forecasting \cite{han2019review}, anomaly detection \cite{chalapathy2019deep}, classification \cite{fawaz2019deep, le2016data}, and so on. There are many data augmentation approaches for time-series data, which can be broadly grouped under two categories \cite{wen2020time}. The first category consists of transformations in time, frequency, and time-frequency domains \cite{wen2020time}, \cite{park2019specaugment}. The second group consists of more advanced methods like decomposition \cite{kegel2018feature}, model-based \cite{cao2014parsimonious}, and learning-based methods \cite{esteban2017real}, \cite{wen2020time}.

Time-domain augmentation methods are straightforward and popular. These approaches directly manipulate the original input time-series data.
For example, the original data is transformed directly by injecting Gaussian noise or other perturbations such as  step-like trend and spikes. Window cropping or sloping also has been used in time domain transformation, which is similar to computer vision method of cropping samples \cite{cui2015data}. 
\an{Other transformations include window warping that compresses or extends a randomly chosen time range and flipping the signal in time-domain}. Additionally, one can use blurring and perturbations in the data points, especially for anomaly detection applications \cite{gao2020robusttad}.
A few approaches have focused on data augmentation in the frequency domain. Gao \textit{et al.} \cite{gao2020robusttad} proposed perturbations for data augmentation in frequency domain, which improves the performance of anomaly detection by convolutional neural networks. The performance of classification was found to be improved by amplitude adjusted Fourier transform and iterated amplitude adjusted Fourier transform which are transformation methods in frequency domain \cite{eileen2019surrogate}.
Time-frequency augmentation methods have also been recenlty investigated. SpecAugment is a Fourier-transform based method that transforms in Mel-Frequency for speech time-series data \cite{park2019specaugment}. The method was found to improve the performance of speech recognition.
In \cite{steven2018feature}, a short Fourier transform is proposed to generate a spectrogram for classification by LSTM neural network.

Decomposition-based, model-based, and learning-based methods are used as advanced data augmentation methods. For decomposition, time-series data are disintegrated to create new data \cite{kegel2018feature}.
Kegel \textit{et al.} firstly decomposes the time-series based on trend, seasonality, and residual. Then, finally new time-series data are generated with a deterministic and a stochastic component.
Bootstrapping methods on the decomposed residuals for generating augmented data was found to help the performance of a forecasting model \cite{bergmeir2016bagging}.
Model-based approaches are related to modeling the dynamics, using statistical model \cite{cao2014parsimonious}, mixture models \cite{kang2020gratis}, and so on. In \cite{cao2014parsimonious}, model-based method were used to address class imbalance for time-series classification. 
Learning-based methods are implemented with learning frameworks such as generative adversarial nets (GAN) \cite{esteban2017real} and reinforcement learning \cite{cubuk2019autoaugment}. These methods generate augmented data by pre-trained models and aim to create realistic synthetic data \cite{esteban2017real}, \cite{cubuk2019autoaugment}.

Finally, augmentation methods can be combined together and applied simultaneously to the data. Combining augmentation methods in time-domain helps to improve performance in classification \cite{um2017data}. However, combining various augmentation methods may results in a large amount of augmented data, increasing training-time, and may not always improve the performance \cite{wen2020time}.

\section{Strategies for Knowledge Distillation with Data Augmentation} 

We would like to investigate strategies for training KD with time-series data and identify augmentation methods for teachers and students that can provide better performance. 
The strategies include two scenarios on KD.
Firstly, we apply augmentation methods only when a student model is trained based on KD with a teacher model trained by the original data. 
Secondly, augmentation methods are applied not only to students, but also to teacher. When a teacher model is trained from scratch, an augmentation method is used, where the model is to be used as a pre-trained model for distillation. And, when a student is trained on KD, the same/different augmentation methods are used.
The set of augmentation approaches on KD are illustrated in Fig. \ref{figure:KD}, and described in further detail later in this section. Also, we explore the effects of ESKD on time-series data -- ESKD uses a teacher which is obtained in the early training process. ESKD  generates better students rather than using the fully-trained teachers from Full KD \cite{cho2019efficacy}. The strategy is derived from the fact that the accuracy is improved initially. However, the accuracy towards the end of training begins to decrease, which is lower than the earlier accuracy. We adopt early stopped teachers with augmentation methods for our experiments presented in Section \ref{Sec:Expts}.

\begin{figure}[htb!]
\includegraphics[scale=0.45] {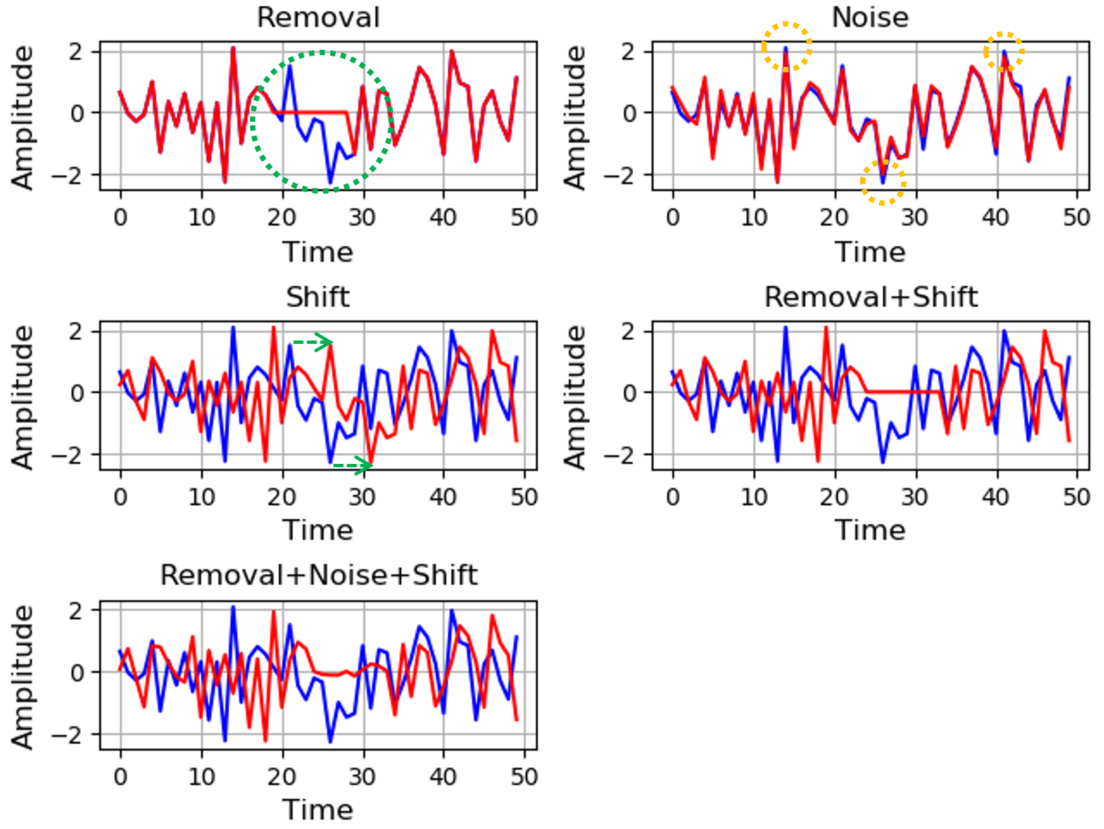} 
\centering
\caption{Illustration of different augmentation  methods used in our knowledge distillation framework. The original data is shown in blue and the corresponding transformed data with data augmentation method is shown in red.}
\label{figure:Aug}
\end{figure}

In order to see effects of augmentation on distillation, we adopt time-domain augmentation methods which are removal, adding noise with Gaussian noise, and shifting. The original pattern, length of the window, and periodical points can be preserved by this transformation. We use transformation methods in time domain so that we can analyze the results from each method, and combinations, more easily. These methods also have been used popularly for training deep learning networks \cite{wen2020time}. We apply combinations of augmentation methods, combined with removal and shifting, and with all methods to a data to see the relationships between each property of datasets for teachers and students of KD. An example of different transformation used for data augmentation is shown in Fig. \ref{figure:Aug}. We describe each of the transforms below:

\begin{itemize}
\item \textbf{Removal:} is used to erase amplitude values of sequential samples. The values of chosen samples to be erased are transformed to the amplitude of the first point. For example, we assume that $n$ samples are chosen as $({X_{t+1}}, {X_{t+2}}, \cdots, {X_{t+n}})$ and their amplitudes 
are $({A_{t+1}}, {A_{t+2}}, \cdots, {A_{t+n}})$ to be erased. ${A_{t+1}}$ is the amplitude of the first sample ${X_{t+1}}$ and is assigned to $(A_{t+1}, A_{t+2}, \cdots, {A_{t+n}}$). That is, values $(A_{t+1}, A_{t+2}, \cdots, A_{t+n})$ are mapped to $(A_{t+1}, A_{t+1}, \cdots, A_{t+1})$. The first point and the number of samples to be erased are chosen randomly. The result of removal is shown in Fig. \ref{figure:Aug} with a green dashed circle.

\item \textbf{Noise Injection:} To inject noise, we apply Gaussian noise with mean 0 and a random standard deviation.  The result of adding noise is shown in Fig. \ref{figure:Aug} with yellow dashed circles. 

\item \textbf{Shifting: } For shifting data, to keep the characteristics such as values of peak points and periodic patterns in the signal, we adopt index shifting and rolling methods to the data for generating new patterns, which means the 100\% shifted signal from the original signal by this augmentation corresponds to the original one. For example, assuming the total number of samples are 50 and 10 time-steps ($20 \%$ of the total number of samples) are chosen to be shifted. The values for amplitude of samples $({X_{1}}, {X_{2}}, \cdots, {X_{11}}, \cdots {X_{50}})$ are $({A_{1}}, {A_{2}}, \cdots, {A_{11}}, \cdots, {A_{49}}, {A_{50}})$. By shifting 10 time-steps, $({A_{41}}, {A_{42}}, \cdots, {A_{1}}, \cdots, {A_{39}}, {A_{40}})$ are newly assigned to the samples of $({X_{1}}, {X_{2}}, \cdots, {X_{11}}, \cdots, {X_{49}}, {X_{50}})$. The number of time-steps to be shifted is chosen randomly. Shifting is shown in Fig. \ref{figure:Aug} with green dashed arrows.

\item \textbf{Mix1: } Applies removal as well as shifting to the same data.

\item \textbf{Mix2: } Applies removal, Gaussian noise injection, and shifting simultaneously to the data. 
\end{itemize}

\section{Experiments and Analysis}\label{Sec:Expts}
In this section, we describe datasets, settings, ablations, and results of our experiments. 

\vspace{-0.1cm}

\subsection{Dataset Description} \label{subsection:Dataset}
We perform experiments on two datasets: GENEActiv \cite{wang2016statistical} and PAMAP2 \cite{reiss2012introducing}, both of which are wearable sensors based activity datasets. We evaluate multiple teachers and students of various capacities for KD with data augmentation methods.

\subsubsection{GENEactiv}
\begin{figure}[h!]
\includegraphics[scale=0.52] {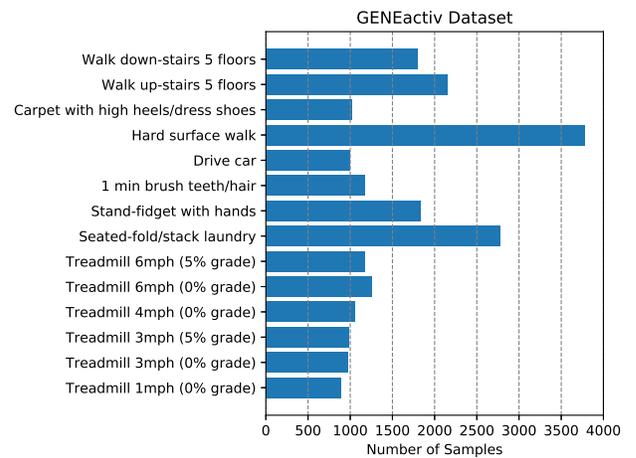}
\caption{Distribution of GENEactive data across different activities. Each sample has 500 time-steps.}
\label{figure:GENEactiv_Dataset}
\end{figure}

GENEactiv dataset \cite{wang2016statistical} consists of 29 activities over 150 subjects. The dataset was collected with a GENEactiv sensor which is a light-weight, waterproof, and wrist-worn tri-axial accelerometer. The sampling frequency of the sensors is 100Hz.
In our experiments, we used 14 activities which can be categorized as daily activities such as walking, sitting, standing, driving, and so on. Each class has over approximately 900 data samples and the distribution and details for activities are illustrated in Fig. \ref{figure:GENEactiv_Dataset}. We split the dataset for training and testing with no overlap in subjects. The number of subjects for training and testing are over 130 and 43, respectively. A window size for a sliding window is 500 time-steps or 5 seconds and the process for temporal windows is full-non-overlapping sliding windows. The number of windows for training is approximately 16000 and testing is 6000.
\vspace{-0.1cm}
\subsubsection{PAMAP2}
\begin{table*}[t]
\caption{Details of PAMAP2 dataset. The dataset consists of 12 activities recorded for 9 subjects.}
\label{table:PAMAP2_Dataset}
\centering
\begin{tabular}{|c|c c c c c c c c c c c|}
\hline
\centering
 & Sbj.101 & Sbj.102 &
Sbj.103 & Sbj.104 & Sbj.105 & Sbj.106 & 
Sbj.107 & Sbj.108 & Sbj.109 & Sum & Nr. of subjects\\
\hline
    Lying & 407 & 350 & 329 & 344 & 354 & 349 & 383 & 361 & 0 & 2877 & 8\\ 
    Sitting & 352 & 335 & 432 & 381 & 402 & 345 & 181 & 342 & 0 & 2770 & 8\\ 
    Standing & 325 & 383 & 307 & 370 & 330 & 365 & 385 & 377 & 0 & 2842 & 8\\ 
    Walking & 333 & 488 & 435 & 479 & 481 & 385 & 506 & 474 & 0 & 3481 & 8\\ 
    Running & 318 & 135 & 0 & 0 & 369 & 341 & 52 & 246 & 0 & 1461 & 6\\ 
    Cycling & 352 & 376 & 0 & 339 & 368 & 306 & 339 & 382 & 0 & 2462 & 7\\
    Nordic walking & 302 & 446 & 0 & 412 & 394 & 400 & 430 & 433 & 0 & 2817 & 7\\ 
    Ascending stairs & 233 & 253 & 147 & 243 & 207 & 192 & 258 & 168 & 0 & 1701 & 8\\ 
    Descending stairs & 217 & 221 & 218 & 206 & 185 & 162 & 167 & 137 & 0 & 1513 & 8\\ 
    Vacuum cleaning & 343 & 309 & 304 & 299 & 366 & 315 & 322 & 364 & 0 & 2622 & 8\\ 
    Ironing & 353 & 866 & 420 & 374 & 496 & 568 & 442 & 496 & 0 & 3995 & 8\\ 
    Rope jumping & 191 & 196 & 0 & 0 & 113 & 0 & 0 & 129 & 92 & 721 & 6\\ 
\hline
\end{tabular}
\end{table*}

PAMAP2 dataset \cite{reiss2012introducing} consists of 18 physical activities for 9 subjects. The 18 activities are categorized as 12 daily activities and 6 optional activities. The dataset was obtained by measurements of heart rate, temperature, accelerometers, gyroscopes, and magnetometers. The sensors were placed on  hands, chest, and ankles of the subject. The total number of dimensions in the time-series is $54$ and the sampling frequency is 100Hz.
To compare with previous methods, in experiments on this dataset, we used leave-one-subject-out combination for validation comparing the $i^{th}$ subject with the $i^{th}$ fold. The input data is in the form of time-series from 40 channels of 4 IMUs and 12 daily activities. To compare with previous methods, the recordings of 4 IMUs are downsampled to 33.3Hz. The 12 action classes are: lying, sitting, standing, walking, running, cycling, nordic walking, ascending stairs, descending stairs, vacuum cleaning, ironing, and rope jumping. Each class and subject are described in Table \ref{table:PAMAP2_Dataset}. There is missing data for some subjects and the distribution of the dataset is imbalanced. A window size for a sliding window is 100 time-steps or 3 seconds and step size is 22 time-steps or 660 ms for segmenting the sequences, which allows semi-non-overlapping sliding windows with 78\% overlapping \cite{reiss2012introducing}.

\subsection{Analysis of Distillation} 
 For experiments on GENEactiv, we run 200 epochs for each model using SGD with momentum 0.9 and the initial learning rate \emph{lr} = 0.1. The \emph{lr} drops by 0.5 after 10 epochs and drops down by 0.1 every [${t \over 3}$] where $t$ is the total number of epochs.
For experiments on PAMAP2, we run 180 epochs for each model using SGD with momentum 0.9 and the initial learning rate \emph{lr} = 0.05. The \emph{lr} drops down by 0.2 after $10$ epochs and drops down $0.1$ every [${t \over 3}$] where $t$ is the total number of epochs.
The results are averaged over 3 runs for both the datasets. To improve the performance, feature engineering \cite{zheng2018feature, bengio2013representation}, feature selection, and reducing confusion by combining classes \cite{dutta2016learning} can be applied additionally. However, to focus on the effects of KD which is based on feature-learning \cite{bengio2013representation}, feature engineering/selection methods to boost performance are not applied and all classes as specified in Section \ref{subsection:Dataset} are used in the following experiments.

\subsubsection{Training from scratch to find a Teacher}

\begin{table*}[h]
\caption{Accuracy for various models trained from scratch on GENEactiv}
\label{table:GENE1}
\centering
\begin{tabular}{ |c c c|c c c|c c c| }
\hline
\centering
Model & $\#$ Parameters & Accuracy ($\%$) &
Model & $\#$ Parameters & Accuracy ($\%$) &
Model & $\#$ Parameters & Accuracy ($\%$)\\
\hline
    ResNet18(8)  & 62,182 & 63.75$\pm$0.42 & WRN16-1 & 61,374 & 67.66$\pm$0.37 & WRN28-1 & 126,782 & 68.63$\pm$0.48 \\ 
    ResNet18(16)  & 244,158 & 65.84$\pm$0.69 & WRN16-2 & 240,318 & 67.84$\pm$0.36 & - & - & -\\ 
    ResNet18(24)  & 545,942 & 66.47$\pm$0.21 & WRN16-3 & 536,254 & 68.89$\pm$0.56 & WRN28-2 & 500,158 & 69.15$\pm$0.24\\ 
    ResNet18(32)  & 967,534 & 66.33$\pm$0.12 & WRN16-4 & 949,438 & 69.00$\pm$0.22 & WRN28-3 & 1,119,550 & 69.23$\pm$0.27\\ 
    ResNet18(48)  & 2,170,142 & 68.13$\pm$0.22 & WRN16-6 & 2,127,550 & 70.04$\pm$0.05 & WRN28-4 & 1,985,214 & 69.29$\pm$0.51\\ 
    ResNet18(64)  & 3,851,982 & 68.17$\pm$0.21 & WRN16-8 & 3,774,654 & 69.02$\pm$0.15 & WRN28-6 & 4,455,358 & 70.99$\pm$0.44\\     
\hline
\end{tabular}
\end{table*}

To find a teacher for KD, we conducted experiments with training from scratch based on two different network architectures: ResNet \cite{he2016deep} and WideResNet \cite{zagoruyko2016wide}. These networks have been popularly used in various state-of-the-art studies for KD \cite{yim2017gift}, \cite{heo2019knowledge}, \cite{goldblum2020adversarially}, \cite{cho2019efficacy}. We modified and compared the structure having the similar number of trainable parameters. As described in Table \ref{table:GENE1}, for training from scratch, WideResNet (WRN) tends to show better performance than ResNet18($k$) where $k$ is the dimension of output from the first layer. The increase in accuracy with the dimension of each block is similar to the basic ResNet.

\subsubsection{Setting hyperparameters for KD}

\begin{figure}
\includegraphics[scale=0.55] {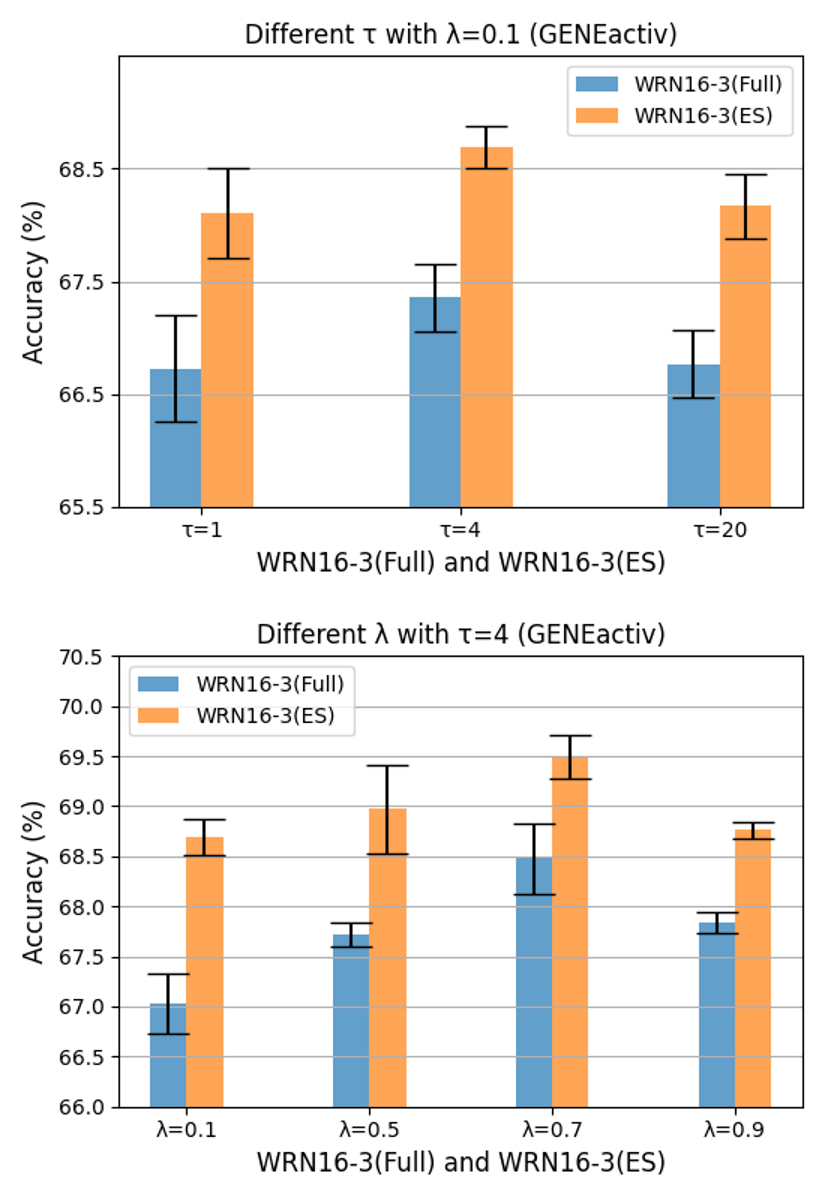}
\centering
\caption{Effect of hyperparmeters $\tau$ and $\lambda$ on the performance of Full KD and ESKD approaches. The results are reported on GENEactive dataset with WRN16-3 and WRN16-1 networks for teacher and student models respectively.}
\label{figure:Temp_Lambda}
\end{figure}

For setting hyperparameters in KD, we conducted several experiments with different temperature $\tau$ as well as lambda $\lambda$. We investigated distillation with different hyperparameters as well. We set WRN16-3 as a teacher network \cite{cho2019efficacy} and WRN16-1 as a student network, which is shown in Fig. \ref{figure:Temp_Lambda}. For temperature $\tau$, in general, $\tau$ $\in\{3,4,5\}$ are used \cite{cho2019efficacy}. High temperature mitigated the peakiness of teachers and helped to make the signal to be softened. In our experiments, according to the results from different $\tau$, high temperature did not effectively help to increase the accuracy. When we used $\tau$ = 4, the results were better than other choices for both datasets with Full KD and ESKD \cite{cho2019efficacy}. For $\lambda$ = 0.7 and 0.99, we obtained the best results with Full KD and ESKD for GENEactiv and PAMAP2, respectively.

\subsubsection{Analyzing Distillation with different size of Models}

\begin{figure}[]
\includegraphics[scale=0.5] {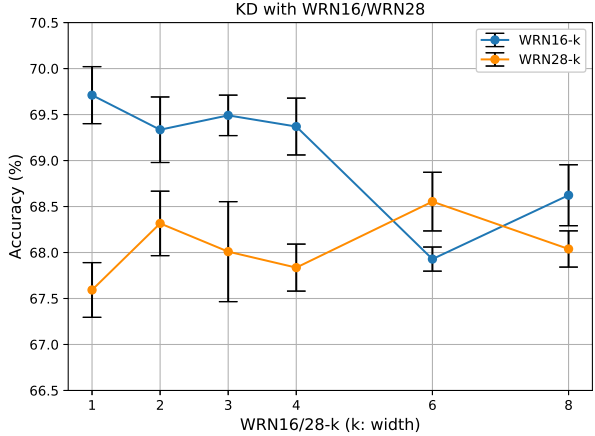}
\centering
\caption{Results of distillation from different teacher models of WRN16-$k$ and WRN28-$k$ on GENEactiv dataset. The higher capacity of teachers does not always increase the accuracy of students.}
\label{figure:GENEactiv_WRN16_WRN28}
\end{figure}

To analyze distillation with different size of models, WRN16-$k$ and WRN28-$k$ were used as teacher networks having different capacity and structures in depth and width $k$. WRN16-1 and WRN28-1 were used as student networks, respectively. As mentioned in the previous section, in general, a higher capacity network trained from scratch shows better accuracy for WRN16 and WRN28. However, as shown in Fig. \ref{figure:GENEactiv_WRN16_WRN28}, in most of the cases, the results from WRN16-$k$ shows better than the results of WRN28-$k$ which has larger width. And the accuracy with teachers of WRN16-3 is higher than the one with teachers having larger width. Therefore, a teacher of higher capacity is not always guaranteed to generate a student whose accuracy is better.

\subsubsection{Knowledge Distillation based on Fully Iterated and Early Stopped Models}

\begin{table}[]
\caption{Accuracy for various models on GENEactiv dataset}
\label{table:GENE_Acc}
\centering
\begin{tabular}{|c c c c|}
\hline
\centering
Student & Teacher & Teacher Acc. (\%) & Student Acc. (\%) \\ \hline
\multirow{6}{5em}{WRN16-1 (ESKD)} & WRN16-2   &  69.06  & 69.34$\pm$0.36            \\
& WRN16-3   &  69.99  & \textbf{69.49}$\pm$0.22            \\
& WRN16-4   &  69.80  & 69.37$\pm$0.31            \\
& WRN16-6   &  70.24  & 67.93$\pm$0.13            \\ 
& WRN16-8   &  70.19  & 68.62$\pm$0.33            \\ \hline
\multirow{2}{5em}{WRN16-1 (Full KD)} & WRN16-3 & 69.68 & 68.62$\pm$0.22  \\ 
& WRN16-8   &  69.28  & 68.68$\pm$0.17            \\ \hline
\end{tabular}
\end{table}

\begin{table}
\caption{Accuracy for various models on PAMAP2 dataset}
\label{table:PAMAP2_1}
\centering
\begin{tabular}{|c c c c|}
\hline
\centering
Student & Teacher & Teacher Acc. (\%) & Student Acc. (\%) \\ \hline
\multirow{6}{5em}{WRN16-1 (ESKD)} & WRN16-2 &  84.86 & 86.18$\pm$2.44 \\
& WRN16-3   &  85.67  & \textbf{86.38}$\pm$2.25             \\
& WRN16-4   &  85.23  & 85.95$\pm$2.27             \\
& WRN16-6   &  85.51  & 86.37$\pm$2.35             \\ 
& WRN16-8   &  85.17  & 85.11$\pm$2.46             \\ \hline
\multirow{2}{5em}{WRN16-1 (Full KD)} & WRN16-3   &  81.52  &  84.31$\pm$2.24          \\
& WRN16-8   &  81.69  &  83.70$\pm$2.52             \\ \hline
\end{tabular}
\end{table}

\begin{table}
\caption{Accuracy ($\%$) for related methods on GENEactiv dataset with 7 classes}\label{table:GENE_Acc_1}
\centering
\begin{tabular}{|p{9em} c c|}
\hline
\centering
\multirow{2}{5em}{Method} & \multicolumn{2}{c|}{Window length} \\
     & 1000 & 500 \\ \hline
    WRN16-1 & 89.29$\pm$0.32 & 86.83$\pm$0.15 \\
    WRN16-3 & 89.53$\pm$0.15 & 87.95$\pm$0.25 \\
    WRN16-8 & 89.31$\pm$0.21 & 87.29$\pm$0.17 \\\hline
    \multirow{2}{9em}{ESKD (WRN16-3)} & \textbf{89.88}$\pm$0.07 & \textbf{88.16}$\pm$0.15 \\
    & (89.74) & (88.30) \\
    \multirow{2}{9em}{ESKD (WRN16-8)} & 89.58$\pm$0.13 & 87.47$\pm$0.11  \\
    & (89.68) & (87.75) \\
    \multirow{2}{9em}{Full KD (WRN16-3)} & 89.84$\pm$0.21 & 87.05$\pm$0.19 \\
    & (88.95) & (86.02) \\
    \multirow{2}{9em}{Full KD (WRN16-8)} & 89.36$\pm$0.06 & 86.38$\pm$0.06  \\
    & (88.74) & (85.08) \\
     \hline
    SVM \cite{cortes1995support} & 86.29 & 85.86  \\
    Choi \textit{et al.} \cite{choi2018temporal} & 89.43 & 87.86 \\ \hline
\end{tabular}
\end{table}

\begin{table}
\caption{Accuracy for related methods on PAMAP2 dataset}
\label{table:PAMAP2_1previous}
\centering
\begin{tabular}{|p{9em} | c |} 
\hline
\centering
Method & Accuracy (\%)\\ \hline

WRN16-1 & 82.81$\pm$2.51  \\
WRN16-3 & 84.18$\pm$2.28  \\
WRN16-8 & 83.39$\pm$2.26  \\ \hline
\multirow{2}{9em}{ESKD (WRN16-3)} & \textbf{86.38}$\pm$2.25 \\
& (85.67) \\
\multirow{2}{9em}{ESKD (WRN16-8)} & 85.11$\pm$2.46  \\
& (85.17) \\
\multirow{2}{9em}{Full KD (WRN16-3)} & 84.31$\pm$2.24 \\
& (81.52) \\
\multirow{2}{9em}{Full KD (WRN16-8)} & 83.70$\pm$2.52  \\
& (81.69) \\ \hline
Chen and Xue \cite{chen2015deep}  &  83.06  \\
Ha \textit{et al.}\cite{ha2015multi}  &  73.79  \\
Ha and Choi \cite{ha2016convolutional}  &  74.21  \\
Kwapisz \cite{kwapisz2011activity}  &  71.27  \\
Catal \textit{et al.} \cite{catal2015use}  &  85.25  \\
Kim \textit{et al.}\cite{kim2012analysis}  &  81.57  \\ \hline
\end{tabular}
\end{table}

We performed additional experiments with WRN16-$k$ which gives the best results. Table \ref{table:GENE_Acc} and Table \ref{table:PAMAP2_1} give detailed results for GENEactiv and PAMAP2, respectively. Compared to training from scratch, although the student capacity from KD is much lower, the accuracy is higher. For instance, for the result of GENEactiv with WRN16-8 by training from scratch, the accuracy is 69.02$\%$ and the number of trainable parameters is 3 million in Table \ref{table:GENE_Acc}. The number of parameters for WRN16-1 as a student for KD is $61$ thousand which is approximately $1.6\%$ of 3 million. However, the accuracy of a student with WRN16-2 teacher from ESKD is $69.34\%$ which is higher than the result of training from scratch with WRN16-8. It shows a model can be compressed with conserved or improved accuracy by KD. Also, we tested with $7$ classes on GENEactiv dataset which were used by the method in \cite{choi2018temporal}. This work used over 50 subjects for testing set. 
Students of KD were WRN16-$1$ and trained with $ \tau = 4$ and $ \lambda = 0.7 $. As shown in Table \ref{table:GENE_Acc_1} where brackets denote the structure of teachers and their accuracy, ESKD from WRN16-3 teacher shows the best accuracy for 7 classes, which is higher than results of models trained from scratch, Full KD, and previous methods \cite{cortes1995support, choi2018temporal}. In most of the cases, students are even better than their teacher. In  various sets of GENEactiv having different number of classes and window length, ESKD shows better performance than Full KD. In Table \ref{table:PAMAP2_1}, the best accuracy on PAMAP2 is $ 86.38\%$ from ESKD with teacher of WRN16-3, which is higher than results from Full KD. The result is even better than previous methods \cite{jordao2018human}, which are described in Table \ref{table:PAMAP2_1previous} where brackets denote the structure of teachers and their accuracy. Therefore, KD allows model compression and improves the accuracy across datasets. And ESKD tends to show better performance compared to Full KD. Also, the higher capacity models as teachers does not always generate better performing student models.

\subsection{Effect of augmentation on student model training} 
To understand distillation effects based on the various capacity of teachers and augmentation methods, WRN16-1, WRN16-3, and WRN16-8 are selected as ``Small'', ``Medium'', and ``Large'' models, respectively. ESKD is used for this experiment which tends to show better performance than the Full KD and requires three-fourths of the total number of epochs for training \cite{cho2019efficacy}.

In order to find augmentation methods impacting KD on students for training, we first trained a teacher from scratch with the original datasets. Secondly, we trained students from the pre-trained teacher with augmentation methods which have different properties including removal, adding noise, shifting, Mix1, and Mix2.
For experiments on GENEactiv, for removal, the number of samples to be removed is less than 50\% of the total number of samples. The first point and the exact number of samples to be erased are chosen randomly. To add noise, the value for standard deviation of Gaussian noise is chosen uniformly at random between 0 and 0.2. For shifting, the number of time-steps to be shifted is less than 50\% of the total number of samples. For Mix1 and Mix2, the same parameters are applied.
For experiments on PAMAP2, the number of samples for removal is less than 10\% of the total number of samples and standard deviation  of Gaussian noise for adding noise is less than 0.1. The parameter for shifting is less than 50\% of the total number of samples. The same parameters of each method are applied for Mix1 and Mix2. 
The length of the window for PAMAP2 is only 100 which is 3 seconds and downsampled from 100Hz data. Compared to GENEactiv whose window size is 500 time-steps or 5 seconds, for PAMAP2, a small transformation can affect the result very prominently. Therefore, lower values are applied to PAMAP2. The parameters for these augmentation methods and the sensor data for PAMAP2 to be transformed are randomly chosen. These conditions for applying augmentation methods are used in the following experiments as well.

\subsubsection{Analyzing augmentation methods on training from scratch and KD}

The accuracy of training scratch with different augmentation methods on WRN16-1 is presented in Table \ref{table:scratch_learning}. Most of the accuracies from augmentation methods, except adding noise which can alter peaky points and change gradients, are higher than the accuracy obtained by learning with the original data. Compared to other methods, adding noise may influence classification between similar activities such as walking, which is included in both datasets as  detailed sub-categories.


\begin{table}[]
\caption{Accuracy ($\%$) of training from scratch on WRN16-1 with different augmentation methods}
\label{table:scratch_learning}
\centering
\begin{tabular}{|p{6em} |c c|}
\hline
\multirow{2}{3em}{Method} & \multicolumn{2}{c|}{Dataset} \\
 & GENEactiv & PAMAP2 \\ \hline
Original & 68.60$\pm$0.23 & 82.81$\pm$2.51 \\
Removal & 69.20$\pm$0.32 & 83.34$\pm$2.41 \\
Noise & 67.60$\pm$0.36 & 82.80$\pm$2.66  \\
Shift & 68.69$\pm$0.22 & 83.91$\pm$2.18 \\
Mix1(R+S) & 69.31$\pm$0.96 & 83.59$\pm$2.37 \\
Mix2(R+N+S) & 67.89$\pm$0.11 & 83.64$\pm$2.76 \\ \hline
\end{tabular}
\end{table}

\begin{figure}[]
\includegraphics[scale=0.56] {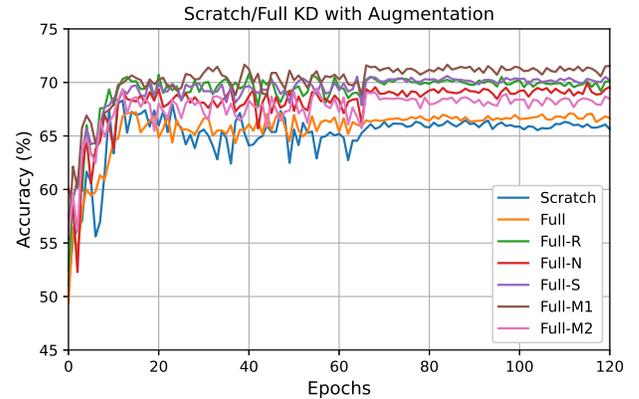}
\centering
\caption{The validation accuracy for training from scratch and Full KD. WRN16-1 is used for training from scratch. For Full KD, WRN16-3 is a teacher network and WRN16-1 is a student network. R, N, S, M1, and M2 in the legend are removal, adding noise, shifting, Mix1, and Mix2, respectively.}
\label{figure:GENEactiv_Scratch_KD}
\end{figure}

The validation accuracy of scratch and Full KD learning on GENEactiv dataset is presented in Fig. \ref{figure:GENEactiv_Scratch_KD}. Training from scratch with the original data shows higher accuracy than KD with original data in very early stages before 25 epochs. However, KD shows better accuracy than the models trained from scratch after 40 epochs. KD with augmentation tends to perform better in accuracy than models trained from scratch and KD learning with the original data alone. That is, data augmentation can help to boost the generalization ability of student models for KD. Mix1 shows the highest accuracy among the results. The highest accuracies are seen in early stages, which are less than 120 epochs for all methods, where 120 epochs is less than three-fourths of the total number of epochs. On closer inspection, we find that the best accuracies are actually seen in less than 20 epochs for training from scratch and Full KD, less than 60 epochs for shifting, Mix1, and Mix2, and less than 120 epochs for adding noise, respectively. This implies that not only early stopped teachers but also early stopped students are able to perform better than fully iterated models. In training based on KD with augmentation methods, the accuracy goes up in early stages, however, the accuracy suffers towards to the end of training. These trends on KD are similar to the previous ESKD study \cite{cho2019efficacy}. For the following experiments, we restrict our analyses to ESKD.

\subsubsection{Analyzing Augmentation Methods on Distillation}

\begin{table}[]
\caption{Accuracy ($\%$) of KD from variants of teacher capacity and augmentation methods on GENEactiv ($\lambda$ = 0.7)}
\label{table:GENEactiv_acc_lambda0.7}
\centering
\begin{tabular}{|p{6em} |p{5em} p{5em} p{5em}|}
\hline
\multirow{3}{3em}{Method} & \multicolumn{3}{c|}{Teacher} \\
 & Small & Medium & Large \\
 & 68.87  & 69.99 & 70.19 \\ \hline
Original & \cellcolor{softy}69.71$\pm$0.31 & 69.61$\pm$0.17 & 68.62$\pm$0.33 \\
Removal & 69.80$\pm$0.34 & 70.23$\pm$0.41 & \cellcolor{softy}70.28$\pm$0.68  \\
Noise & 69.26$\pm$0.08 & \cellcolor{softy}69.12$\pm$0.19 & 69.38$\pm$0.39 \\
Shift & \cellcolor{softy}70.63$\pm$0.19 & 70.43$\pm$0.89 & 70.00$\pm$0.20 \\
Mix1(R+S) & 70.56$\pm$0.57 & \cellcolor{softy}\textbf{71.35}$\pm$0.20 & 70.22$\pm$0.10 \\
Mix2(R+N+S) & 69.27$\pm$0.31 & 69.51$\pm$0.28 & \cellcolor{softy}69.62$\pm$0.21 \\ \hline
\end{tabular}
\end{table}

\begin{table}[]
\caption{Accuracy ($\%$) of KD from variants of teacher capacity and augmentation methods on GENEactiv ($\lambda$ = 0.99)}
\label{table:GENEactiv_acc_lambda0.99}
\centering
\begin{tabular}{|p{6em} |p{5em} p{5em} p{5em}|}
\hline
\multirow{3}{3em}{Method} & \multicolumn{3}{c|}{Teacher} \\
 & Small & Medium & Large \\
 & 68.87  & 69.99 & 70.19 \\ \hline
Original & \cellcolor{softy}69.44$\pm$0.19 & 67.80$\pm$0.36 & 68.67$\pm$0.20 \\
Removal & 69.48$\pm$0.22 & 69.75$\pm$0.40 & \cellcolor{softy}70.01$\pm$0.27  \\
Noise & 69.99$\pm$0.14 & \cellcolor{softy}70.20$\pm$0.06 & 70.12$\pm$0.14 \\
Shift & \cellcolor{softy}\textbf{70.96}$\pm$0.10 & 70.42$\pm$0.06 & 70.16$\pm$0.24 \\
Mix1(R+S) & \cellcolor{softy}70.40$\pm$0.27 & 70.07$\pm$0.38 & 69.36$\pm$0.16 \\
Mix2(R+N+S) & \cellcolor{softy}70.56$\pm$0.23 & 69.88$\pm$0.16 & 69.71$\pm$0.30 \\ \hline
\end{tabular}
\end{table}

\begin{table}[]
\caption{Accuracy ($\%$) of KD from variants of teacher capacity and augmentation methods on PAMAP2 ($\lambda$ = 0.7)}
\label{table:PAMAP2_acc_lambda0.7}
\centering
\begin{tabular}{|p{6em} |p{5em} p{5em} p{5em}|}
\hline
\multirow{3}{3em}{Method} & \multicolumn{3}{c|}{Teacher} \\
 & Small & Medium & Large \\
 & 85.42  & 85.67 & 85.17 \\ \hline
Original & 84.75$\pm$2.64 & 84.47$\pm$2.32 &  \cellcolor{softy}84.90$\pm$2.38 \\
Removal & 85.16$\pm$2.46 & \cellcolor{softy}85.51$\pm$2.27 & 85.02$\pm$2.47  \\
Noise & 84.96$\pm$2.59 & \cellcolor{softy}85.52$\pm$2.26 & 84.85$\pm$2.43 \\
Shift & 85.21$\pm$2.21 & 85.45$\pm$2.19 &  \cellcolor{softy}\textbf{85.66}$\pm$2.26 \\
Mix1(R+S) & 85.54$\pm$2.51 & \cellcolor{softy}85.60$\pm$2.19 & 84.71$\pm$2.53 \\
Mix2(R+N+S) & 85.17$\pm$2.39 & \cellcolor{softy}85.27$\pm$2.33 & 83.76$\pm$2.77 \\ \hline
\end{tabular}
\end{table}

\begin{table}[h!]
\caption{Accuracy ($\%$) of KD from variants of teacher capacity and augmentation methods on PAMAP2 ($\lambda$ = 0.99)}
\label{table:PAMAP2_acc_lambda0.99}
\centering
\begin{tabular}{|p{6em} |p{5em} p{5em} p{5em}|}
\hline
\multirow{3}{3em}{Method} & \multicolumn{3}{c|}{Teacher} \\
 & Small & Medium & Large \\
 & 85.42  & 85.67 & 85.17 \\ \hline
Original & 86.37$\pm$2.35 &  \cellcolor{softy}86.38$\pm$2.25 & 85.11$\pm$2.46 \\
Removal & 84.66$\pm$2.67 &  \cellcolor{softy}85.70$\pm$2.40 & 84.81$\pm$2.52  \\
Noise & 84.77$\pm$2.65 &  \cellcolor{softy}85.21$\pm$2.41 & 85.05$\pm$2.40 \\
Shift & 86.08$\pm$2.42 &  \cellcolor{softy}\textbf{86.65}$\pm$2.13 & 85.53$\pm$2.28 \\
Mix1(R+S) & 84.93$\pm$2.71 &  \cellcolor{softy}85.88$\pm$2.28 & 84.73$\pm$2.54 \\
Mix2(R+N+S) & 82.94$\pm$2.76 &  \cellcolor{softy}83.94$\pm$2.70 & 83.28$\pm$2.50 \\ \hline
\end{tabular}
\end{table}

The accuracy of each augmentation method with KD is summarized in Table \ref{table:GENEactiv_acc_lambda0.7} and \ref{table:GENEactiv_acc_lambda0.99} for GENEactiv and Table \ref{table:PAMAP2_acc_lambda0.7} and \ref{table:PAMAP2_acc_lambda0.99} for PAMAP2. The results were obtained from small-sized students of ESKD. The gray colored cells of these tables are the best accuracy for the augmentation method among the different capacity teachers of KD.
When a higher $\lambda$ is used, distillation from teachers is improved, and the best results are obtained when the teacher capacity is smaller. Also,  the best performance of students, when learning with augmentation methods and the original data, is achieved with similar teacher capacities. 
For example, for GENEactiv with $\lambda$ = 0.7, the best results are generated from various capacity of teachers. But, with $\lambda$ = 0.99, the best results tend to be seen with smaller capacity of teachers. Even though the evaluation protocol for PAMAP2 is leave-one-subject-out with an imbalanced distribution of data, with $\lambda$ = 0.7, the best results are obtained from larger capacity of teachers as well. Furthermore, results from both datasets verify that larger and more accurate teachers do not always result in better students. Also, the best result from shifting is seen at the same capacity of the teacher with the original data. It might be because shifting includes the same time-series `shapes' as the original data. The method for shifting is simple but is an effectively helpful method for training KD. For all teachers on PAMAP2 with $\lambda$ = 0.99, the accuracies from training by shifting are even higher than other combinations. Compared to previous methods \cite{jordao2018human} with PAMAP2, the result by shifting outperforms others. Furthermore, although the student network of KD has the same number of parameters of the network trained from scratch (WRN16-1), the accuracy is much higher than the latter one; the result of Mix1 from GENEactiv and shifting from PAMAP2 by the medium teacher is approximately 2.7$\%$ points and 3.8$\%$ points better than the result from original data by training from scratch, respectively. These accuracies are even better than the results of their teachers. It also verifies that KD with an augmentation method including shifting has benefits to obtain improved results.

\begin{table}[h!]
\caption{$p$-value and (accuracy ($\%$), standard deviation) for training from scratch and KD on GENEactiv dataset}
\label{table:GENE_pvalue}
\centering
\begin{tabular}{|l |l c|}

\hline
\centering
\multirow{2}{*}{Scratch} & KD & \multirow{2}{*}{$p$-value} \\
&(Teacher: Medium)&\\\hline
\multirow{14}{6em}{Original (68.60$\pm$0.23)} & Original (ESKD)  & \multirow{2}{*}{0.030} \\
 & (69.61$\pm$0.17) & \\  \cline{2-3}
 &  Original (Full)  & \multirow{2}{*}{0.045}   \\
 & (68.62$\pm$0.22) & \\  \cline{2-3}
 & Removal  &  \multirow{2}{*}{0.006} \\
 & (70.23$\pm$0.41) & \\  \cline{2-3}
 & Noise  & \multirow{2}{*}{0.012} \\
  & (69.12$\pm$0.19) & \\  \cline{2-3}
 & Shift  & \multirow{2}{*}{0.025} \\ 
  & (70.43$\pm$0.89) & \\  \cline{2-3}
 &  Mix1(R+S) & \multirow{2}{*}{0.073}  \\
  & (71.35$\pm$0.20) & \\  \cline{2-3}
 & Mix2(R+N+S) & \multirow{2}{*}{0.055} \\
  & (69.51$\pm$0.28) & \\ \hline

\end{tabular}
\end{table}

\begin{table}[h!]
\caption{$p$-value and (accuracy ($\%$), standard deviation) for training from scratch and KD on PAMAP2 dataset}
\label{table:PAMAP2_pvalue}
\centering

\begin{tabular}{|l |l c|}

\hline
\centering
\multirow{2}{*}{Scratch} & KD & \multirow{2}{*}{$p$-value} \\
&(Teacher: Medium)&\\\hline
\multirow{14}{6.3em}{Original (82.81$\pm$2.51)} & Original (ESKD)  & \multirow{2}{*}{0.0298} \\
 & (84.47$\pm$2.32) & \\  \cline{2-3}
 &  Original (Full)  & \multirow{2}{*}{0.0007}   \\
 & (84.31$\pm$2.24) & \\  \cline{2-3}
 & Removal  &  \multirow{2}{*}{0.0008} \\
 & (85.51$\pm$2.27) & \\  \cline{2-3}
 & Noise  & \multirow{2}{*}{0.0002} \\
  & (85.52$\pm$2.26) & \\  \cline{2-3}
 & Shift  & \multirow{2}{*}{0.0034} \\ 
  & (85.45$\pm$2.19) & \\  \cline{2-3}
 &  Mix1(R+S) & \multirow{2}{*}{0.0024}  \\
  & (85.60$\pm$2.19) & \\  \cline{2-3}
 & Mix2(R+N+S) & \multirow{2}{*}{0.0013} \\
  & (85.27$\pm$2.33) & \\ \hline

\end{tabular}
\end{table}

To investigate the difference in performance with a model trained from scratch and KD with augmentation methods, statistical analysis was conducted by calculating $p$-value from a $t$-test with a confidence level of 95$\%$. Table \ref{table:GENE_pvalue} and \ref{table:PAMAP2_pvalue} show averaged accuracy, standard deviation, and calculated $p$-value for WRN16-1 trained from scratch with original training set and various student models of WRN16-1 trained with KD and augmentation. That is, student models in KD have the same structure of the model trained from scratch and teachers for KD are WRN16-3 ($\tau$ = 4, $\lambda$ = 0.7).
For GENEactiv, in five out of the seven cases, the calculated $p$-values are less than 0.05. Thus, the results in the table show  statistically-significant difference between training from scratch and KD. For PAMAP2, in all cases, $p$-values are less than 0.05. This also represents statistically-significant difference between training from scratch and KD. Therefore, we can conclude that KD training with augmentation methods, which shows better results in classification accuracy, performs significantly different from training from scratch, at a confidence level of 95$\%$.

\begin{table}[h!]
\caption{ECE ($\%$) of training from scratch and KD on GENEactiv dataset }
\label{table:GENE_ece}
\centering
\begin{tabular}{|l c |l c|}
\hline
\centering
\multirow{2}{*}{Scratch} & \multirow{2}{*}{ECE} & KD & \multirow{2}{*}{ECE} \\
&&(Teacher: Medium)&\\
\hline
Original & 3.22 & Original (ESKD)  & 2.96 \\
Removal & 3.56  &  Removal  &  2.90 \\
Noise & 3.45  &  Noise  & 2.85 \\
Shift & 3.24  &  Shift  & 2.78 \\ 
Mix1(R+S) & 3.72  &  Mix1(R+S)  & 2.79  \\
Mix2(R+N+S) & 3.67 & Mix2(R+N+S) & 2.86 \\ \hline
\end{tabular}
\end{table}

\begin{table}[h!]
\caption{ECE ($\%$) of training from scratch and KD on PAMAP2 dataset}
\label{table:PMAAP2_ece}
\centering
\begin{tabular}{|l c |l c|}
\hline
\centering
\multirow{2}{*}{Scratch}  & \multirow{2}{*}{ECE} & KD & \multirow{2}{*}{ECE} \\
&&(Teacher: Medium) &\\\hline
Original & 2.28 & Original (ESKD)  & 2.16 \\
Removal & 3.64 &  Removal  &  3.09 \\
Noise & 5.83 &  Noise  & 3.01  \\
Shift & 2.87 &  Shift  & 2.22  \\ 
Mix1(R+S) & 4.39 &  Mix1(R+S)  & 2.96  \\
Mix2(R+N+S) & 5.55 & Mix2(R+N+S) & 4.17  \\ \hline
\end{tabular}
\end{table}

Finally, the expected calibration error (ECE) \cite{guo2017calibration} is calculated to measure the confidence of performance for models trained from scratch and KD ($\tau$ = 4, $\lambda$ = 0.7) with augmentation methods. As shown in Table \ref{table:GENE_ece} and \ref{table:PMAAP2_ece}, in all cases, ECE values for KD are lower than when models are trained from scratch, indicating that models trained with KD have higher reliability. Also, results of KD including shifting are lower than results from other augmentation methods. This additionally verifies that KD improves the performance and shifting helps to get improved models.

\subsubsection{Analyzing training for KD with augmentation methods}

\begin{table}[h!]
\caption{The loss value (${\mathbf 10}^{-2}$) for KD (teacher: medium) from various methods on GENEactiv}
\label{table:GENE_loss}
\centering
\begin{tabular}{|p{6em} p{5em} p{5em} p{4em}|}
\hline
\centering
Method ($\lambda$=0.7) & CE \par Train & KD \par Train & KD \par Test \\ \hline
Original & 3.774 & 0.617 & 1.478 \\
Removal & 3.340 & 0.406 & 1.246 \\
Noise & 11.687 & \textbf{1.172} & 1.358 \\
Shift & 2.416 & \textbf{0.437} & 1.119 \\
Mix1(R+S) & 5.475 & 0.475 & \textbf{1.108} \\
Mix2(R+N+S) & 17.420 & 1.337 & 1.338 \\ \hline
\end{tabular}
\end{table}

\begin{table}[h!]
\caption{The loss value (${\mathbf 10}^{-2}$) for KD (teacher: medium) from various methods on PAMAP2 (Subject 101)}
\label{table:PAMAP2_loss}
\centering
\begin{tabular}{|p{6em} p{5em} p{5em} p{4em}|}
\hline
\centering

Method ($\lambda$=0.7) & CE \par Train & KD \par Train & KD \par Test \\ \hline
Original & 0.832 & 0.156 & 1.783 \\ 
Removal & 1.237 & 0.146 & 1.038 \\
Noise & 1.066 & \textbf{0.138} & 1.284 \\
Shift & 0.468 & \textbf{0.129} & 1.962 \\
Mix1(R+S) & 1.267 & 0.150 & \textbf{0.895} \\
Mix2(R+N+S) & 1.853 & 0.177 & 1.065 \\ \hline

\end{tabular}
\end{table}

The loss values of each method, for the medium-sized teacher, are shown in Table \ref{table:GENE_loss} and \ref{table:PAMAP2_loss}. The loss values were obtained from the final epoch while training student models based on Full KD. As shown in these tables, for both cross entropy and KD loss values, training with shifting-based data augmentation results in lower loss, compared to other augmentation strategies and the original model. The loss value for noise augmentation is higher than the values of shifting. On the other hand, the KD loss value for Mix1 is higher than the values for removal and shifting. However, the training loss  is for these two methods and its value of testing is lower. Compared to other methods, Mix2 shows higher loss for training, which may be because this method generates more complicated patterns. However, the testing KD loss value of Mix2 is lower than the value of original and adding noise. These findings imply that the data of original and shifting have very similar patterns. And data based on Mix1 and Mix2 are not simply trainable data for distillation, however, these methods have an effect of preventing a student from over-fitting or degradation for classification. The contrast of results from GENEactiv between each method is more prominent than the one from PAMAP2. This is due to the fact that smaller parameters for augmentation are applied to PAMAP2. Also, the dataset is more challenging to train on, due to imbalanced data and different channels in sensor data.

\subsection{Analysis of Teacher and Student Models with a Variant Properties of Training Set}
To discuss properties of training set for teacher and student models, we use the same parameter ($\tau$ = 4, $\lambda$ = 0.7) in this experiment on two datasets. In this section, we try to train a medium teacher and a small student by training set having the same or different properties to take into account relationships between teachers and students. Testing set is not transformed or modified. The medium teacher is chosen because the teacher showed good performance in our prior experiments discussed in previous sections. Further, distillation from a medium model to a small model is an preferable approach \cite{cho2019efficacy}. Also, we analyze which augmentation method is effective to achieve higher accuracy. We use adding noise, shifting, and Mix1 methods which transform data differently.

\begin{table}[h]
\caption{Accuracy ($\%$) of training from scratch on WRN16-3 with different augmentation methods}
\label{table:scratch_learning2}
\centering
\begin{tabular}{|p{4.5em}|p{4.5em} p{4.5em} p{4.5em} p{4.5em}|}
\hline
Dataset & Original & Noise & Shift & Mix1(R+S)  \\ \hline
GENEactiv & 69.53$\pm$0.40 & 68.59$\pm$0.05 & 72.08$\pm$0.20 & 71.64$\pm$0.26 \\

GENEactiv & \multirow{2}{3em}{69.99} &
 \multirow{2}{3em}{68.68} & \multirow{2}{3em}{72.48} &
 \multirow{2}{3em}{72.17} \\
(Top-1) &&&&\\ \hline
PAMAP2 & 84.65$\pm$2.28 & 83.08$\pm$2.51 & 82.54$\pm$2.42 & 82.39$\pm$2.62 \\
PAMAP2 & \multirow{2}{3em}{85.67} &
 \multirow{2}{3em}{85.31} & \multirow{2}{3em}{84.38} &
 \multirow{2}{3em}{84.09} \\
(Top-1) &&&&\\
 \hline
\end{tabular}
\end{table}

To obtain a medium teacher model, the model is trained from scratch with augmentation methods. These results are shown in Table \ref{table:scratch_learning2}. For GENEactiv, shifting based data augmentation gives the best performance. However, for PAMAP2, original data achieves the best performance. Mix1 shows slightly lower accuracy than shifting. In these experiments, the student model is trained using the teacher model that achieves best performance over several trials.

\begin{figure}
\includegraphics[scale=0.55] {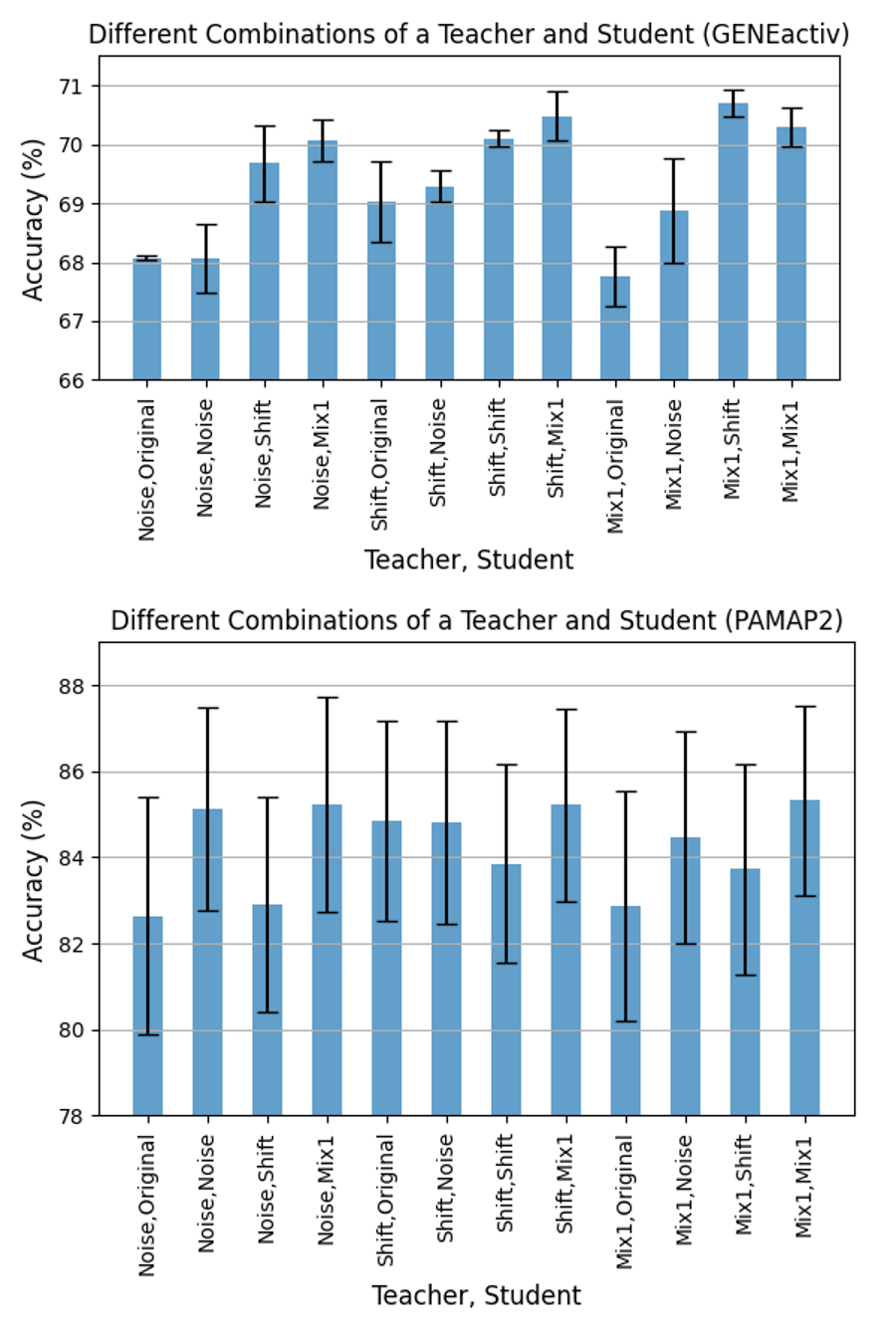}
\centering
\caption{The results for students trained by different combinations of training sets for teachers and students. The teacher and student both are learned by augmentation methods. WRN16-3 (medium) and WRN16-1 (small) are teacher and student networks, respectively.}
\label{figure:TeacherStudent}
\end{figure}

We also evaluated different combinations of data augmentation strategies for teacher-student network pairs. A pair is obtained by using one or no data augmentation strategy to train the teacher network by training from scratch, and the student network is trained by ESKD under different, same, or no augmentation strategy. The results  are shown in Fig. \ref{figure:TeacherStudent}. We found that KD with the same data augmentation strategy for training teachers and students may not be the right choice to get the best performance.
When a teacher is trained by shifting and a student is trained by Mix1 which showed good performance as a student in the previous sections, the results are better than other combinations for both datasets. Also, when a student is learned by Mix1 including shifting transform, in general, the performance are also good for all teachers.
It implies that the method chosen for training a student is more important than choosing a teacher; KD with a medium teacher trained by the original data and a student trained with shift or Mix1 outperforms other combinations. Using the same strategy for training data for teachers and students does not always present the best performance. When the training set for students is more complicated than the set for teachers, the performance in accuracy tends to be better. That is, applying a transformation method to students can help to increase the accuracy.
It also verifies that better teachers do not always lead to increased accuracy of students.
Even if the accuracies from these combinations of a teacher and student are lower than models trained from scratch by WRN16-3, the number of parameters for the student is only about 11\% of the one for WRN16-3. Therefore, the results still are good when considering both performance and computation.

\subsection{Analysis of Student Models with Different Data Augmentation Strategies for Training and Testing Set}
In this section, we study the effect of students on KD from various augmentation methods for training and testing, while a teacher is trained with the original dataset. We use the same parameter ($\tau$ = 4, $\lambda$ = 0.7) and ESKD for this experiment on two datasets. A teacher is selected with a medium model trained by the original data. We use adding noise, shifting and Mix1 methods which transform data differently.

\begin{figure}[]
\includegraphics[scale=0.53] {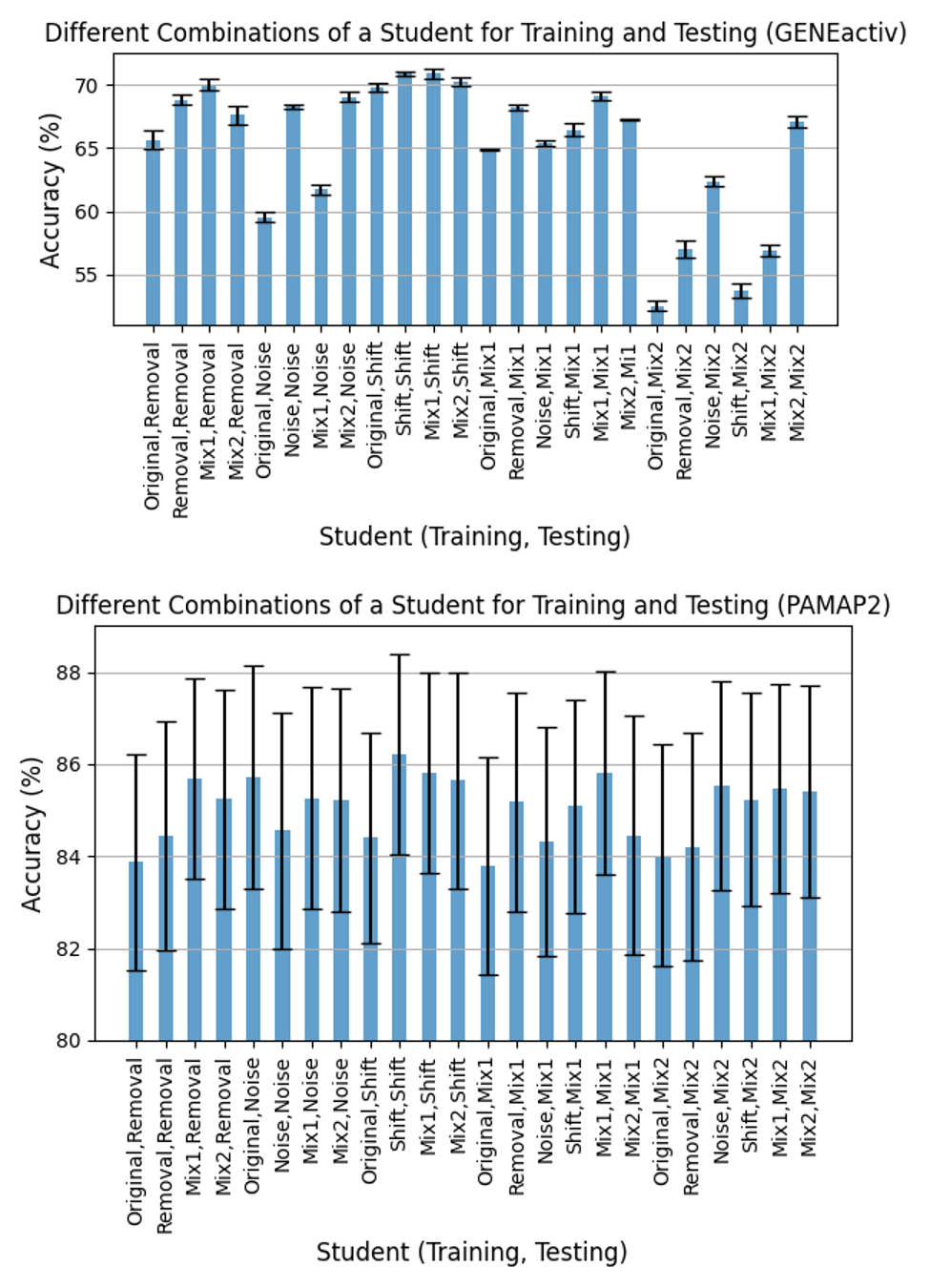}
\centering
\caption{Effect on classification performance of student network with different augmentation methods for training and testing sets. WRN16-3 (medium) and WRN16-1 (small) are teacher and student networks, respectively.}
\label{figure:StudentAug}
\end{figure}
After training the teacher network on original data, a student network is trained with different data augmentation strategies and is evaluated on test data transformed with different data augmentation strategies. The results are illustrated in Fig. \ref{figure:StudentAug}. For GENEactiv, most often, training student networks with Mix1 show better performance on different testing sets. However, if the testing set is affected by adding noise, training students with adding noise and Mix2 shows much better performance than training with shifting and Mix1. From the results on PAMAP2, in most of the cases, training students with Mix1 shows better performance to many different testing set. However, when the testing set is augmented by adding noise, training with original data shows the best performance. This is likely attributable to the window size, which has about a hundred samples, and the dataset includes the information of 4 kinds of IMUs. Therefore, injecting noise, which can affect peaky points and change gradients, creates difficulties for classification. Also, these issue can affect the both training and testing data. Thus, if the target data includes noise, training set and augmentation methods have to be considered along with the length of the window and intricate signal shapes within the windows.

\subsection{Analysis of Testing Time} 
Here, we compare the evaluation time for various models on the GENEactiv dataset. We conducted the test on a desktop with a 3.50 GHz CPU (Intel® Xeon(R) CPU E5-1650 v3), 48 GB memory, and NVIDIA TITAN Xp (3840 NVIDIA® CUDA® cores and 12 GB memory) graphic card. We used a batch size of 1 and approximately $6000$ data samples for testing. Four different models were trained from scratch with WRN16-$k$ ($k$=1, 3, 6, and 8). To test with ESKD and Mix1, WRN16-3 was used as a teacher and WRN16-1 was used for student network. As expected, larger models take more time for testing, as shown in Table \ref{table:processing_time2}. WRN16-1 as a student trained by ESKD with Mix1 augmentation achieves the best accuracy, 71.35$\%$, where the model takes the least amount of time on both GPU and CPU. 
The results on CPU reiterate the reason why model compression is required for many applications, especially on edge devices, wearables,  and mobile devices, which have limited computational and power resources and are generally implemented in real time with only CPU. The gap in performance would be higher if an edge device had lower computational resources. 

\begin{table}[h]
\caption{Processing time of various models for GENEactive dataset}
\label{table:processing_time2}
\centering
\begin{tabular}{|p{7.9em}|p{2.5em} p{3em} p{2.5em} p{3em} p{3em}|} 
\hline
\multirow{3}{7.9em}{Model (WRN16-$k$)} &
 \multirow{3}{3em}{Acc. (\%)} & Total & Avg. & Total & Avg.\\
 && GPU & GPU & CPU & CPU\\
 && (sec) & (ms) & (sec) & (ms)\\ \hline

$k$=1 & 67.66 & \multirow{3}{*}{15.226} & \multirow{3}{*}{2.6644} & \multirow{3}{*}{\textbf{16.655}} & \multirow{3}{*}{2.8920} \\
$k$=1 (ESKD) & 69.61 & & & &  \\
$k$=1 (ESKD+Mix1) & \textbf{71.35} & & & & \\ \hline
$k$=3 & 68.89 & 16.426 & 2.8524 & 21.333 & 3.7044 \\
$k$=6 & 70.04 & 16.663 & 2.8934 & 33.409 & 5.8012 \\
$k$=8 & 69.02 & 16.885 & 2.9320 & 46.030 & 7.9928 \\ \hline

\end{tabular}
\end{table}

\section{Conclusion}
In this paper, we studied many relevant aspects of knowledge distillation (KD) for wearable sensor data as applied to human activity analysis. 
We conducted experiments with different sizes of teacher networks to evaluate their effect on KD performance. We show that a high capacity teacher network does not necessarily ensure better performance of a student network. We further showed that training with augmentation methods and early stopping for KD (ESKD) is effective when dealing with time-series data. We also establish that the choice of augmentation strategies has more of an impact on the student network training as opposed to the teacher network. 
In most cases, KD training with the Mix1 (Removal+Shifting) data augmentation strategy for students showed robust performance. 
Further, we also conclude that a single augmentation strategy is not conclusively better all the time. Therefore, we recommend using a combination of augmentation methods for training KD in general. In summary, our findings provide a comprehensive understanding of KD and data augmentation strategies for time-series data from wearable devices of human activity. These conclusions can be used as a general set of recommendations to establish a strong baseline performance on new datasets and new applications.  

\section*{Acknowledgment}

This research was funded by NIH R01GM135927, as part of the Joint DMS/NIGMS Initiative to Support Research at the Interface of the Biological and Mathematical Sciences, and by NSF CAREER grant 1452163.





\ifCLASSOPTIONcaptionsoff
  \newpage
\fi

{\footnotesize
\bibliographystyle{IEEEtran}
\bibliography{main2.bib}

\begin{thebibliography}{10}
\providecommand{\url}[1]{#1}
\csname url@samestyle\endcsname
\providecommand{\newblock}{\relax}
\providecommand{\bibinfo}[2]{#2}
\providecommand{\BIBentrySTDinterwordspacing}{\spaceskip=0pt\relax}
\providecommand{\BIBentryALTinterwordstretchfactor}{4}
\providecommand{\BIBentryALTinterwordspacing}{\spaceskip=\fontdimen2\font plus
\BIBentryALTinterwordstretchfactor\fontdimen3\font minus
  \fontdimen4\font\relax}
\providecommand{\BIBforeignlanguage}[2]{{%
\expandafter\ifx\csname l@#1\endcsname\relax
\typeout{** WARNING: IEEEtran.bst: No hyphenation pattern has been}%
\typeout{** loaded for the language `#1'. Using the pattern for}%
\typeout{** the default language instead.}%
\else
\language=\csname l@#1\endcsname
\fi
#2}}
\providecommand{\BIBdecl}{\relax}
\BIBdecl

\bibitem{he2016deep}
K.~He, X.~Zhang, S.~Ren, and J.~Sun, ``Deep residual learning for image
  recognition,'' in \emph{Proceedings of the IEEE Conference on Computer Vision
  and Pattern Recognition}, 2016, pp. 770--778.

\bibitem{huang2017densely}
G.~Huang, Z.~Liu, L.~Van Der~Maaten, and K.~Q. Weinberger, ``Densely connected
  convolutional networks,'' in \emph{Proceedings of the IEEE Conference on
  Computer Vision and Pattern Recognition}, 2017, pp. 4700--4708.

\bibitem{dalal2005histograms}
N.~Dalal and B.~Triggs, ``Histograms of oriented gradients for human
  detection,'' in \emph{Proceedings of the IEEE Conference on Computer Vision
  and Pattern Recognition}, 2005, pp. 886--893.

\bibitem{lowe2004distinctive}
D.~G. Lowe, ``Distinctive image features from scale-invariant keypoints,''
  \emph{International Journal of Computer Vision}, vol.~60, no.~2, pp. 91--110,
  2004.

\bibitem{abdel2014convolutional}
O.~Abdel-Hamid, A.-r. Mohamed, H.~Jiang, L.~Deng, G.~Penn, and D.~Yu,
  ``Convolutional neural networks for speech recognition,'' \emph{IEEE/ACM
  Transactions on Audio, Speech, and Language Processing}, vol.~22, no.~10, pp.
  1533--1545, 2014.

\bibitem{xiong2018microsoft}
W.~Xiong, L.~Wu, F.~Alleva, J.~Droppo, X.~Huang, and A.~Stolcke, ``The
  microsoft 2017 conversational speech recognition system,'' in
  \emph{Proceedings of the IEEE International Conference on Acoustics, Speech
  and Signal Processing}, 2018, pp. 5934--5938.

\bibitem{wan2020deep}
S.~Wan, L.~Qi, X.~Xu, C.~Tong, and Z.~Gu, ``Deep learning models for real-time
  human activity recognition with smartphones,'' \emph{Mobile Networks and
  Applications}, vol.~25, no.~2, pp. 743--755, 2020.

\bibitem{fawaz2019deep}
H.~I. Fawaz, G.~Forestier, J.~Weber, L.~Idoumghar, and P.-A. Muller, ``Deep
  learning for time series classification: a review,'' \emph{Data Mining and
  Knowledge Discovery}, vol.~33, no.~4, pp. 917--963, 2019.

\bibitem{khan2020survey}
A.~Khan, A.~Sohail, U.~Zahoora, and A.~S. Qureshi, ``A survey of the recent
  architectures of deep convolutional neural networks,'' \emph{Artificial
  Intelligence Review}, vol.~53, no.~8, pp. 5455--5516, 2020.

\bibitem{gil2020improving}
M.~Gil-Mart{\'\i}n, R.~San-Segundo, F.~Fernandez-Martinez, and
  J.~Ferreiros-L{\'o}pez, ``Improving physical activity recognition using a new
  deep learning architecture and post-processing techniques,''
  \emph{Engineering Applications of Artificial Intelligence}, vol.~92, p.
  103679, 2020.

\bibitem{molchanov2016pruning}
P.~Molchanov, S.~Tyree, T.~Karras, T.~Aila, and J.~Kautz, ``Pruning
  convolutional neural networks for resource efficient inference,'' in
  \emph{Proceedings of the International Conference on Learning
  Representations}, 2017.

\bibitem{han2015deep}
S.~Han, H.~Mao, and W.~J. Dally, ``Deep compression: Compressing deep neural
  networks with pruning, trained quantization and huffman coding,'' in
  \emph{Proceedings of the International Conference on Learning
  Representations}, 2016.

\bibitem{wu2016quantized}
J.~Wu, C.~Leng, Y.~Wang, Q.~Hu, and J.~Cheng, ``Quantized convolutional neural
  networks for mobile devices,'' in \emph{Proceedings of the IEEE Conference on
  Computer Vision and Pattern Recognition}, 2016, pp. 4820--4828.

\bibitem{tai2015convolutional}
C.~Tai, T.~Xiao, Y.~Zhang, X.~Wang \emph{et~al.}, ``Convolutional neural
  networks with low-rank regularization,'' in \emph{Proceedings of the
  International Conference on Learning Representations}, 2016.

\bibitem{hinton2015distilling}
G.~Hinton, O.~Vinyals, and J.~Dean, ``Distilling the knowledge in a neural
  network,'' in \emph{Proceedings of the International Conference on Neural
  Information Processing Systems Deep Learning and Representation Learning
  Workshop}, 2015.

\bibitem{yim2017gift}
J.~Yim, D.~Joo, J.~Bae, and J.~Kim, ``A gift from knowledge distillation: Fast
  optimization, network minimization and transfer learning,'' in
  \emph{Proceedings of the IEEE Conference on Computer Vision and Pattern
  Recognition}, 2017, pp. 4133--4141.

\bibitem{heo2019knowledge}
B.~Heo, M.~Lee, S.~Yun, and J.~Y. Choi, ``Knowledge distillation with
  adversarial samples supporting decision boundary,'' in \emph{Proceedings of
  the AAAI Conference on Artificial Intelligence}, vol.~33, 2019, pp.
  3771--3778.

\bibitem{cho2019efficacy}
J.~H. Cho and B.~Hariharan, ``On the efficacy of knowledge distillation,'' in
  \emph{Proceedings of the IEEE International Conference on Computer Vision},
  2019, pp. 4794--4802.

\bibitem{bucilua2006model}
C.~Buciluǎ, R.~Caruana, and A.~Niculescu-Mizil, ``Model compression,'' in
  \emph{Proceedings of the ACM SIGKDD International Conference on Knowledge
  Discovery and Data Mining}, 2006, pp. 535--541.

\bibitem{zagoruyko2016paying}
S.~Zagoruyko and N.~Komodakis, ``Paying more attention to attention: Improving
  the performance of convolutional neural networks via attention transfer,'' in
  \emph{Proceedings of the International Conference on Learning
  Representations}, 2017.

\bibitem{tung2019similarity}
F.~Tung and G.~Mori, ``Similarity-preserving knowledge distillation,'' in
  \emph{Proceedings of the IEEE International Conference on Computer Vision},
  2019, pp. 1365--1374.

\bibitem{tarvainen2017mean}
A.~Tarvainen and H.~Valpola, ``Mean teachers are better role models:
  Weight-averaged consistency targets improve semi-supervised deep learning
  results,'' in \emph{Proceedings of the International Conference on Neural
  Information Processing Systems}, 2017, pp. 1195--1204.

\bibitem{zhang2018deep}
Y.~Zhang, T.~Xiang, T.~M. Hospedales, and H.~Lu, ``Deep mutual learning,'' in
  \emph{Proceedings of the IEEE Conference on Computer Vision and Pattern
  Recognition}, 2018, pp. 4320--4328.

\bibitem{goldblum2020adversarially}
M.~Goldblum, L.~Fowl, S.~Feizi, and T.~Goldstein, ``Adversarially robust
  distillation,'' in \emph{Proceedings of the AAAI Conference on Artificial
  Intelligence}, vol.~34, no.~04, 2020, pp. 3996--4003.

\bibitem{furlanello2018born}
T.~Furlanello, Z.~Lipton, M.~Tschannen, L.~Itti, and A.~Anandkumar, ``Born
  again neural networks,'' in \emph{Proceedings of the International Conference
  on Machine Learning}, 2018, pp. 1607--1616.

\bibitem{yang2019training}
C.~Yang, L.~Xie, S.~Qiao, and A.~L. Yuille, ``Training deep neural networks in
  generations: A more tolerant teacher educates better students,'' in
  \emph{Proceedings of the AAAI Conference on Artificial Intelligence},
  vol.~33, 2019, pp. 5628--5635.

\bibitem{han2019review}
Z.~Han, J.~Zhao, H.~Leung, K.~F. Ma, and W.~Wang, ``A review of deep learning
  models for time series prediction,'' \emph{IEEE Sensors Journal}, vol.~21,
  no.~6, 2019.

\bibitem{chalapathy2019deep}
R.~Chalapathy and S.~Chawla, ``Deep learning for anomaly detection: A survey,''
  \emph{arXiv preprint arXiv:1901.03407}, 2019.

\bibitem{le2016data}
A.~Le~Guennec, S.~Malinowski, and R.~Tavenard, ``Data augmentation for time
  series classification using convolutional neural networks,'' in
  \emph{ECML/PKDD workshop on advanced analytics and learning on temporal
  data}, 2016.

\bibitem{wen2020time}
Q.~Wen, L.~Sun, X.~Song, J.~Gao, X.~Wang, and H.~Xu, ``Time series data
  augmentation for deep learning: A survey,'' \emph{arXiv preprint
  arXiv:2002.12478}, 2020.

\bibitem{park2019specaugment}
D.~S. Park, W.~Chan, Y.~Zhang, C.-C. Chiu, B.~Zoph, E.~D. Cubuk, and Q.~V. Le,
  ``Specaugment: A simple data augmentation method for automatic speech
  recognition,'' in \emph{Proceedings of the Interspeech}, 2019, pp.
  2613--2617.

\bibitem{kegel2018feature}
L.~Kegel, M.~Hahmann, and W.~Lehner, ``Feature-based comparison and generation
  of time series,'' in \emph{Proceedings of the International Conference on
  Scientific and Statistical Database Management}, 2018, pp. 1--12.

\bibitem{cao2014parsimonious}
H.~Cao, V.~Y. Tan, and J.~Z. Pang, ``A parsimonious mixture of gaussian trees
  model for oversampling in imbalanced and multimodal time-series
  classification,'' \emph{IEEE Transactions on Neural Networks and Learning
  Systems}, vol.~25, no.~12, pp. 2226--2239, 2014.

\bibitem{esteban2017real}
C.~Esteban, S.~L. Hyland, and G.~R{\"a}tsch, ``Real-valued (medical) time
  series generation with recurrent conditional gans,'' \emph{arXiv preprint
  arXiv:1706.02633}, 2017.

\bibitem{cui2015data}
X.~Cui, V.~Goel, and B.~Kingsbury, ``Data augmentation for deep neural network
  acoustic modeling,'' \emph{IEEE/ACM Transactions on Audio, Speech, and
  Language Processing}, vol.~23, no.~9, pp. 1469--1477, 2015.

\bibitem{gao2020robusttad}
J.~Gao, X.~Song, Q.~Wen, P.~Wang, L.~Sun, and H.~Xu, ``Robusttad: Robust time
  series anomaly detection via decomposition and convolutional neural
  networks,'' \emph{arXiv preprint arXiv:2002.09545}, 2020.

\bibitem{eileen2019surrogate}
K.~T.~L. Eileen, Y.~Kuah, K.-H. Leo, S.~Sanei, E.~Chew, and L.~Zhao,
  ``Surrogate rehabilitative time series data for image-based deep learning,''
  in \emph{Proceedings of the European Signal Processing Conference}, 2019, pp.
  1--5.

\bibitem{steven2018feature}
O.~Steven~Eyobu and D.~S. Han, ``Feature representation and data augmentation
  for human activity classification based on wearable imu sensor data using a
  deep lstm neural network,'' \emph{Sensors}, vol.~18, no.~9, p. 2892, 2018.

\bibitem{bergmeir2016bagging}
C.~Bergmeir, R.~J. Hyndman, and J.~M. Ben{\'\i}tez, ``Bagging exponential
  smoothing methods using stl decomposition and box--cox transformation,''
  \emph{International journal of forecasting}, vol.~32, no.~2, pp. 303--312,
  2016.

\bibitem{kang2020gratis}
Y.~Kang, R.~J. Hyndman, and F.~Li, ``Gratis: Generating time series with
  diverse and controllable characteristics,'' \emph{Statistical Analysis and
  Data Mining: The ASA Data Science Journal}, vol.~13, no.~4, pp. 354--376,
  2020.

\bibitem{cubuk2019autoaugment}
E.~D. Cubuk, B.~Zoph, D.~Mane, V.~Vasudevan, and Q.~V. Le, ``Autoaugment:
  Learning augmentation strategies from data,'' in \emph{Proceedings of the
  IEEE Conference on Computer Vision and Pattern Recognition}, 2019, pp.
  113--123.

\bibitem{um2017data}
T.~T. Um, F.~M. Pfister, D.~Pichler, S.~Endo, M.~Lang, S.~Hirche, U.~Fietzek,
  and D.~Kuli{\'c}, ``Data augmentation of wearable sensor data for
  parkinson’s disease monitoring using convolutional neural networks,'' in
  \emph{Proceedings of the 19th ACM International Conference on Multimodal
  Interaction}, 2017, pp. 216--220.

\bibitem{wang2016statistical}
Q.~Wang, S.~Lohit, M.~J. Toledo, M.~P. Buman, and P.~Turaga, ``A statistical
  estimation framework for energy expenditure of physical activities from a
  wrist-worn accelerometer,'' in \emph{Proceedings of the Annual International
  Conference of the IEEE Engineering in Medicine and Biology Society}, vol.
  2016, 2016, pp. 2631--2635.

\bibitem{reiss2012introducing}
A.~Reiss and D.~Stricker, ``Introducing a new benchmarked dataset for activity
  monitoring,'' in \emph{Proceedings of the International Symposium on Wearable
  Computers}, 2012, pp. 108--109.

\bibitem{zheng2018feature}
A.~Zheng and A.~Casari, \emph{Feature engineering for machine learning:
  principles and techniques for data scientists}.\hskip 1em plus 0.5em minus
  0.4em\relax O'Reilly Media, Inc., 2018.

\bibitem{bengio2013representation}
Y.~Bengio, A.~Courville, and P.~Vincent, ``Representation learning: A review
  and new perspectives,'' \emph{IEEE transactions on pattern analysis and
  machine intelligence}, vol.~35, no.~8, pp. 1798--1828, 2013.

\bibitem{dutta2016learning}
A.~Dutta, O.~Ma, M.~P. Buman, and D.~W. Bliss, ``Learning approach for
  classification of geneactiv accelerometer data for unique activity
  identification,'' in \emph{2016 IEEE 13th International Conference on
  Wearable and Implantable Body Sensor Networks (BSN)}.\hskip 1em plus 0.5em
  minus 0.4em\relax IEEE, 2016, pp. 359--364.

\bibitem{zagoruyko2016wide}
S.~Zagoruyko and N.~Komodakis, ``Wide residual networks,'' in \emph{Proceedings
  of the British Machine Vision Conference}, 2016.

\bibitem{cortes1995support}
C.~Cortes and V.~Vapnik, ``Support-vector networks,'' \emph{Machine learning},
  vol.~20, no.~3, pp. 273--297, 1995.

\bibitem{choi2018temporal}
H.~Choi, Q.~Wang, M.~Toledo, P.~Turaga, M.~Buman, and A.~Srivastava, ``Temporal
  alignment improves feature quality: an experiment on activity recognition
  with accelerometer data,'' in \emph{Proceedings of the IEEE Conference on
  Computer Vision and Pattern Recognition Workshops}, 2018, pp. 349--357.

\bibitem{chen2015deep}
Y.~Chen and Y.~Xue, ``A deep learning approach to human activity recognition
  based on single accelerometer,'' in \emph{Proceedings of the IEEE
  International Conference on Systems, Man, and Cybernetics}, 2015, pp.
  1488--1492.

\bibitem{ha2015multi}
S.~Ha, J.-M. Yun, and S.~Choi, ``Multi-modal convolutional neural networks for
  activity recognition,'' in \emph{Proceedings of the IEEE International
  Conference on Systems, Man, and Cybernetics}, 2015, pp. 3017--3022.

\bibitem{ha2016convolutional}
S.~Ha and S.~Choi, ``Convolutional neural networks for human activity
  recognition using multiple accelerometer and gyroscope sensors,'' in
  \emph{Proceedings of the International Joint Conference on Neural Networks},
  2016, pp. 381--388.

\bibitem{kwapisz2011activity}
J.~R. Kwapisz, G.~M. Weiss, and S.~A. Moore, ``Activity recognition using cell
  phone accelerometers,'' \emph{ACM SigKDD Explorations Newsletter}, vol.~12,
  no.~2, pp. 74--82, 2011.

\bibitem{catal2015use}
C.~Catal, S.~Tufekci, E.~Pirmit, and G.~Kocabag, ``On the use of ensemble of
  classifiers for accelerometer-based activity recognition,'' \emph{Applied
  Soft Computing}, vol.~37, pp. 1018--1022, 2015.

\bibitem{kim2012analysis}
H.-J. Kim, M.~Kim, S.-J. Lee, and Y.~S. Choi, ``An analysis of eating
  activities for automatic food type recognition,'' in \emph{Proceedings of the
  Asia Pacific Signal and Information Processing Association Annual Summit and
  Conference}, 2012, pp. 1--5.

\bibitem{jordao2018human}
A.~Jordao, A.~C. Nazare~Jr, J.~Sena, and W.~R. Schwartz, ``Human activity
  recognition based on wearable sensor data: A standardization of the
  state-of-the-art,'' \emph{arXiv preprint arXiv:1806.05226}, 2018.

\bibitem{guo2017calibration}
C.~Guo, G.~Pleiss, Y.~Sun, and K.~Q. Weinberger, ``On calibration of modern
  neural networks,'' in \emph{Proceedings of the International Conference on
  Machine Learning}, 2017, pp. 1321--1330.

\end{thebibliography}
}

%


\begin{IEEEbiography}[{\includegraphics[width=1in,height=1.25in,clip,keepaspectratio]{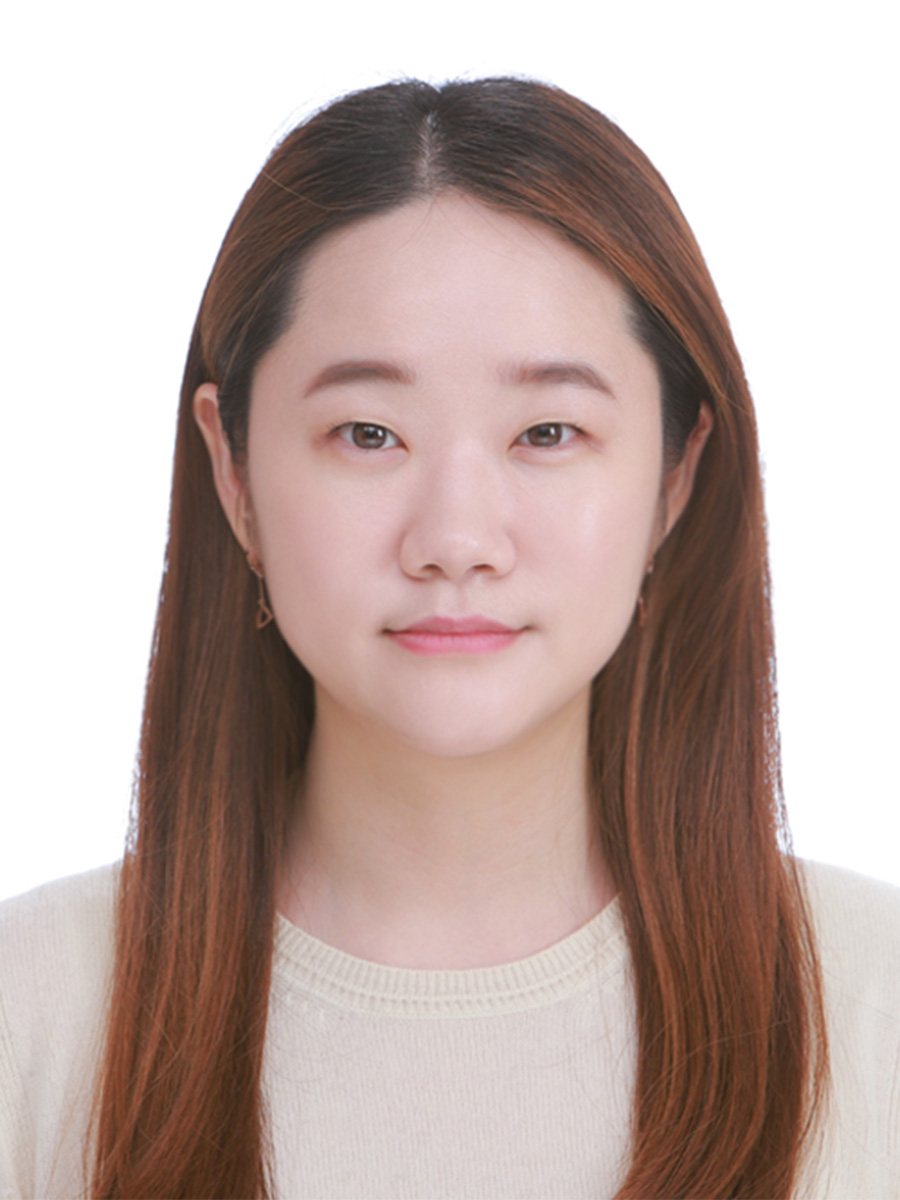}}]{Eun Som Jeon}
received the B.E. and M.E. degrees in Electronics and Electrical Engineering from Dongguk University, Seoul, Korea in 2014 and 2016, respectively. She worked in Korea Telecom (Institute of Convergence Technology), Seoul, Korea.
She is currently pursuing the Ph.D. degree in Computer Engineering (Electrical Engineering) with Geometric Media Laboratory, Arizona State University, Tempe, AZ, USA. 
Her current research interests include time-series and image data analysis, human behavior analysis, deep learning, and artificial analysis.
\end{IEEEbiography}

\begin{IEEEbiography}[{\includegraphics[width=1in,height=1.25in,clip,keepaspectratio]{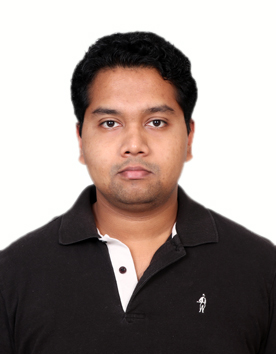}}]{Anirudh Som}
is an Advanced Computer Scientist in the Center for Vision Technologies group at SRI International. He  received  his M.S.  and  Ph.D.  degrees  in Electrical Engineering from  Arizona State University in 2016 and 2020 respectively, prior to which he received his B.Tech. degree in Electronics and Communication Engineering from GITAM University in India. His research interests are in the fields of machine learning, computer vision, human movement analysis, human behavior analysis and dynamical system analysis.
\end{IEEEbiography}

\begin{IEEEbiography}[{\includegraphics[width=1in,height=1.25in,clip,keepaspectratio]{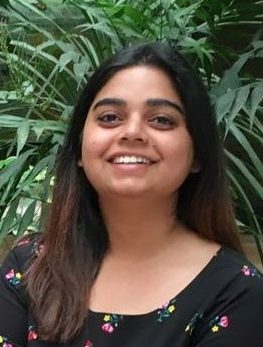}}]{Ankita Shukla}
is a postdoctoral researcher at Arizona State University. She received her PhD and Masters degrees in Electronics and Communication from IIIT-Delhi, India in 2020 and 2014 respectively. Her research interest are in the field of machine learning, computer vision, time-series data analysis and geometric methods.  
\end{IEEEbiography}


\begin{IEEEbiography}[{\includegraphics[width=1in,height=1.25in,clip,keepaspectratio]{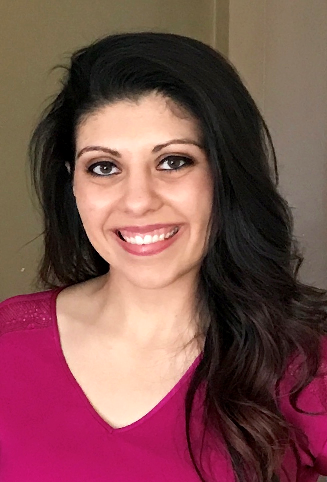}}]{Kristina Hasanaj}
is a graduate research associate at Arizona State University. She earned her B.S. in Exercise Science (Kinesiology concentration) and M.A. in Exercise Physiology from Central Michigan University. She is currently pursuing her doctoral degree through the Nursing and Healthcare Innovation Ph.D. program at Arizona State University. Her research interests are focused around behaviors in the 24-hour day (sleep, sedentary behavior, physical activity) and the use of mobile health and wearable technologies in clinical and health related settings.

\end{IEEEbiography}

\begin{IEEEbiography}[{\includegraphics[width=1in,height=1.25in,clip,keepaspectratio]{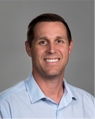}}]{Matthew P. Buman}, PhD is an associate professor in the College of Health Solutions at Arizona State University. His research interests reflect the dynamic interplay of behaviors in the 24-hour day, including sleep, sedentary behavior, and physical activity. His work focuses on the measurement of these behaviors using wearable technologies, interventions that singly or in combination target these behaviors, and the environments that impact these behaviors.

\end{IEEEbiography} 

\begin{IEEEbiography}[{\includegraphics[width=1in,height=1.25in,clip,keepaspectratio]{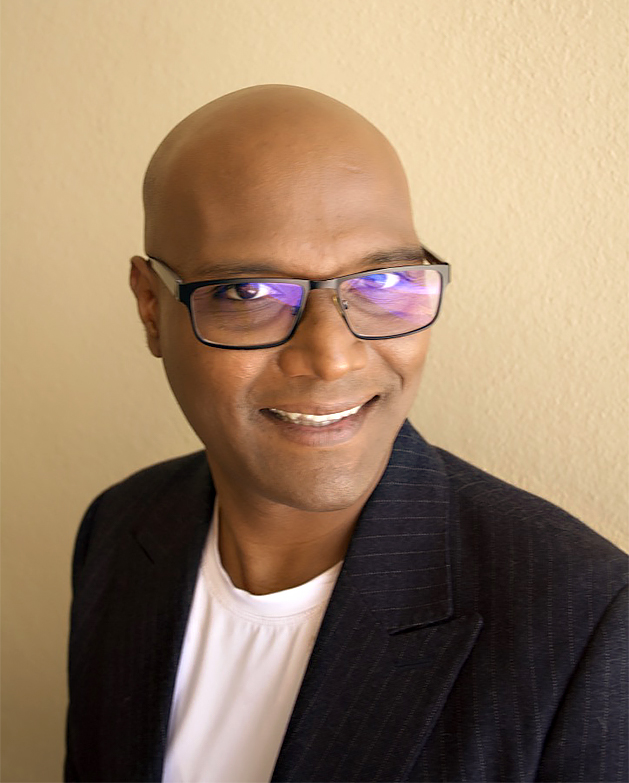}}]{Pavan Turaga}, PhD is an associate professor in the School of Arts, Media and Engineering at Arizona State University. He received a bachelor's degree in electronics and communication engineering from the Indian Institute of Technology Guwahati, India, in 2004, and a master's and doctorate in electrical engineering from the University of Maryland, College Park in 2007 and 2009, respectively. His research interests include computer vision and computational imaging with applications in activity analysis, dynamic scene analysis, and time-series data analysis with geometric methods.

\end{IEEEbiography}




\end{document}